\documentclass[acmtog]{acmart}
\usepackage{makecell}
\usepackage{pifont}
\usepackage{multirow}
\AtBeginDocument{%
  }

\setcopyright{acmcopyright}
\copyrightyear{2018}
\acmYear{2018}
\acmDOI{XXXXXXX.XXXXXXX}

\acmJournal{TOG}
\acmVolume{37}
\acmNumber{4}
\acmArticle{111}
\acmMonth{8}

\begin{document}

\title{HAvatar: High-fidelity Head Avatar via Facial Model Conditioned Neural Radiance Field}
\author{XIAOCHEN ZHAO}
\orcid{0000-0001-8976-7723}
\email{zhaoxc19@mails.tsinghua.edu.cn}
\affiliation{%
  \institution{Tsinghua University}
  \city{Beijing}
  \country{China}
}

\author{LIZHEN WANG}
\orcid{0000-0002-6674-9327}
\email{wang-lz@mail.tsinghua.edu.cn}
\affiliation{%
  \institution{Tsinghua University \& NNKosmos Technology}
  \city{Beijing \& Hangzhou}
  \country{China}
}

\author{JINGXIANG SUN}
\orcid{0000-0003-2966-9501}
\email{starkjxsun@gmail.com}
\affiliation{%
  \institution{Tsinghua University}
  \city{Beijing}
  \country{China}
}

\author{HONGWEN ZHANG}
\orcid{0000-0002-6674-9327}
\email{zhanghongwen@mail.tsinghua.edu.cn}
\affiliation{%
  \institution{Tsinghua University}
  \city{Beijing}
  \country{China}
}

\author{JINLI SUO}
\orcid{0000-0002-3426-1634}
\email{jlsuo@tsinghua.edu.cn}
\affiliation{%
  \institution{Tsinghua University}
  \city{Beijing}
  \country{China}
}

\author{YEBIN LIU}
\orcid{0000-0003-3215-0225}
\email{liuyebin@mail.tsinghua.edu.cn}
\affiliation{%
  \institution{Tsinghua University}
  \city{Beijing}
  \country{China}
}

\renewcommand{\shortauthors}{Zhao et al.}

\begin{abstract}

The problem of modeling an animatable 3D human head avatar under light-weight setups is of significant importance but has not been well solved. Existing 3D representations either perform well in the realism of portrait images synthesis or the accuracy of expression control, but not both. To address the problem, we introduce a novel hybrid explicit-implicit 3D representation, Facial Model Conditioned Neural Radiance Field, which integrates the expressiveness of NeRF and the prior information from the parametric template. At the core of our representation, a synthetic-renderings-based condition method is proposed to fuse the prior information from the parametric model into the implicit field without constraining its topological flexibility. Besides, based on the hybrid representation, we properly overcome the inconsistent shape issue presented in existing methods and improve the animation stability. Moreover, by adopting an overall GAN-based architecture using an image-to-image translation network, we achieve high-resolution, realistic and view-consistent synthesis of dynamic head appearance. Experiments demonstrate that our method can achieve state-of-the-art performance for 3D head avatar animation compared with previous methods.
\end{abstract}



\begin{teaserfigure}
  \includegraphics[width=\textwidth]{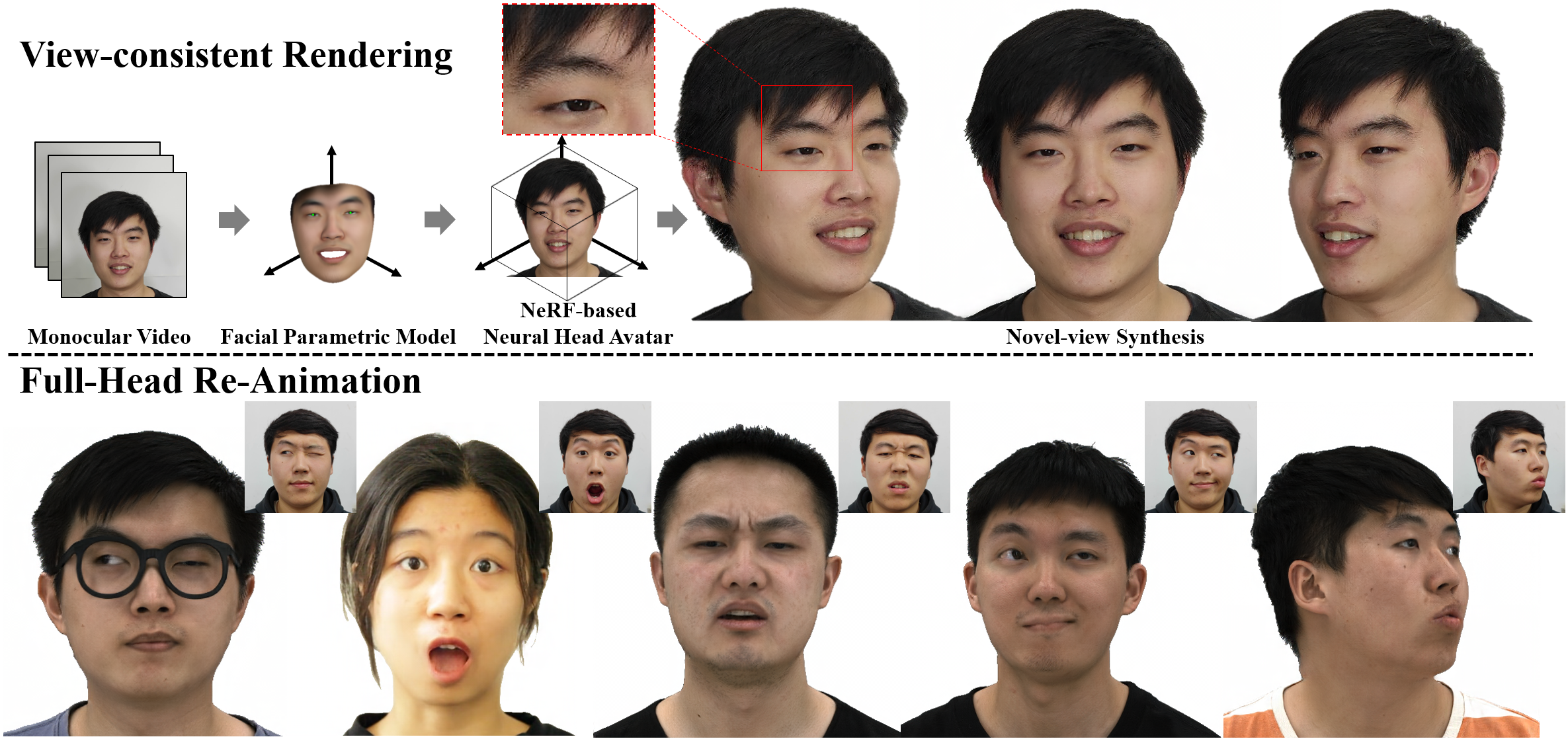}
  \caption{Our method is able to synthesize high-resolution, photo-realistic and view-consistent head images, achieving fine-grained control over head poses and facial expressions.}
  \label{fig:teaser}
\end{teaserfigure}
\maketitle

\section{Introduction}
Animatable 3D human head avatar modeling is of great significance in many applications such as VR/AR games and telepresence. There are two key factors for a lifelike virtual personal character: the accuracy of facial expression control and the realism of portrait images synthesis. 
Though exiting solutions~\cite{lombardi2018deep, lombardi2021mixture, ma2021pixel} are able to reconstruct high-quality dynamic human heads, they typically depend on complicated dense-view capture systems and even rely on hundreds of cameras. By leveraging the learning-based techniques, researchers have shifted interest to explore the possibility of automatically modeling human head avatar, with accurate controllability and high-fidelity appearance, under light-weight setup.

Firstly, to establish a controllable personalized head character, the most straightforward way is to directly learn a global parameter conditioned neural head avatar from image sequences, but such method~\cite{gafni2021nerface} limits the generalization ability in expression control. To improve control robustness, other works~\cite{zheng2021imavatar, grassal2022neural} attempt to leverage parametric templates~\cite{TOG2017flame} to help regulate the avatar modeling during the training stage. However, the explicit surface prior from the parametric model constrains the expressive power for complex-topology parts (i.e. glasses).

Secondly, for high-fidelity human head avatar modeling, recent implicit-surface-based methods~\cite{yenamandra2021i3dmm, zheng2021imavatar, grassal2022neural} recover more texture details compared with conventional methods~\cite{yang2020facescape, TVCG2014FaceWarehouse, TOG2017flame, wang2022faceverse} with limited-resolution texture representation. Nevertheless, the quality of the recovered appearance is still far from satisfactory. Built on the expressive neural radiance field (NeRF)~\cite{mildenhall2020nerf}, Nerface~\cite{gafni2021nerface} is able to generate more promising dynamic appearance results. However, based on the MLP backbone, it is trained in an auto-decoding fashion and tends to overfit training sequences, leading to the obvious inconsistent shape across different frames and unnatural head shaking in the test phase.

Combining the expressiveness of NeRF and the prior information from the parametric template is a promising way for achieving fine-grained expression control and realistic portrait synthesis. Recent work~\cite{athar2022rignerf} establishes a deformable-mesh-guided dynamic NeRF for head avatar modeling.
However, the prominent challenge for the coupling of geometry models and NeRF comes from the difficulty in establishing reliable dense correspondences between the real-world subject and the fitted parametric template. 
Due to the limited expressiveness of the morphable model, it is hard for the deformed mesh to perfectly align with the real-world head with high diversity in terms of geometry and topology. Resulted from the obvious misalignment, the spatial sampling points in the neural radiance field tend to establish ambiguous correspondences with the mesh surface, leading to blurry or unstable rendering performance.

In this paper, we introduce a novel Parametric Model-Conditioned Neural Radiance Field for Human Head Avatar.
Inspired by the effective rendering-to-video translation architecture adopted by~\cite{kim2018dvp}, we extend the synthetic-rendering-based condition for 3D head control, by integrating it with triplane-based neural volumetric representation~\cite{chan2022eg3d}. The dynamic head character is conditioned by the axis-aligned feature planes generated by the orthogonal renderings of the textured fitted parametric model in the canonical space. We leverage a powerful convolutional network to learn the reasonable correspondences between the canonical synthetic renderings and the observed head appearance, hence avoiding the ambiguous correspondences determined by the Euclidean distance. On the one hand, such a synthetic-rendering-based condition introduces the prior of the fully-controllable 3D facial model into the neural representation to achieve fine-grained and consistent expression control. On the other hand, orthogonal renderings can supply rough 3D descriptions and avoid excessive restriction from the coarse geometry of the model mesh, so that our head avatar is capable of describing complex topology structure. Considering that the dynamic content mainly comes from facial expressions, we utilize a facial parametric model rather than a full-head model in practice, leaving only the facial region benefiting from the model‘s prior.

\begin{figure*}
\begin{center}
\includegraphics[scale=0.34]{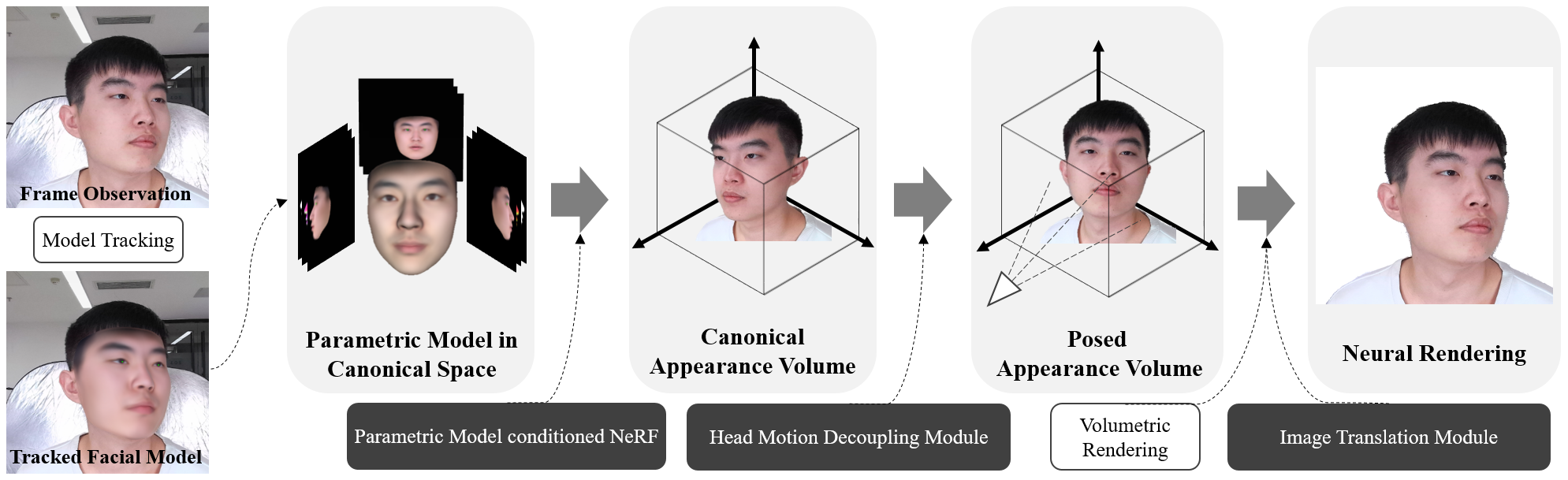}
\end{center}
\caption{The overview of our parametric-model-based Neural Head Avatar.}
\label{fig:pipeline}
\end{figure*}

Moreover, while retaining the powerful appearance expressiveness of NeRF~\cite{mildenhall2020nerf}, our method also overcomes the inconsistent shape issue that commonly occurs in NeRF-based modeling methods~\cite{gafni2021nerface}.
Based on our synthetic-renderings-based orthogonal-plane representation, we utilize learnable embeddings to modulate the plane feature generators rather than condition the MLP decoder in an auto-decoding fashion like Nerface~\cite{gafni2021nerface}.
By modulating the convolutional kernels and normalizing the feature generation, the embeddings are able to regulate the whole feature volume to avoid overfitting, leading to the consistent head shape in the animation.
Our experiments prove that, with per-frame embeddings modulating on the convolutional generators, the shape consistency and the animation stability of our head avatar are significantly improved.

Finally, our method inherits the advantage of NeRF~\cite{mildenhall2020nerf}, which intrinsically supports differentiable rendering and maintains multiview consistency. Thanks to this strength, we further integrate the NeRF-based volume rendering with the neural rendering, and optimize the whole architecture end-to-end with image observations to recover facial details. Specifically, by leveraging the effective image-to-image translation network commonly used in researches of portrait video synthesis~\cite{zakharov2019few, xu2020deep3dportrait, thies2019deferred, chen2020puppeteergan, kim2018dvp}, we translate the rendered 3D-aware feature map into RGB images. Training the overall network in an adversarial manner, our solution firstly achieves high-resolution and view-consistent photo-realistic synthesis for 3D head avatar.

Given monocular or sparse-views videos, after fitting per frame 3D facial models with an off-the-shelf tracker, our approach is able to learn a high-fidelity and view-consistent personalized 3D head avatar, including hair, accessories and torso, under full control of head poses and facial expressions. Meanwhile, we optimize a linear blend skinning (LBS) weight field as well, that decouples the motions of the head and the torso via a backward warping. During test time, 
given a single-view driving video, pose and expression parameters are extracted to deform the facial model, and our method can faithfully recover the entire head appearance under novel expressions, poses, and viewpoints.

In summary, the main contributions of this paper include:
\begin{itemize}
\item We propose a novel facial model conditioned NeRF for personalized 3D head avatar, which is built on an orthogonal synthetic-renderings based feature volume. Our representation enables flexible topology and accurate control over the head motion and facial expressions.
\item Benefiting from our hybrid representation, we develop a new strategy of modulating generators with conditional embeddings to handle the inconsistent shape issue presented in existing NeRF-based avatar modeling methods and significantly improve the animation stability.
\item We firstly achieve high-resolution realistic and view-consistent synthesis of dynamic head appearance, by adopting an overall GAN-based architecture combining our efficient avatar representation with an image-to-image translation module.
\item Besides the learning head avatar from monocular videos, we also present head avatar modeling from multiview videos (using 6 cameras), and experiments demonstrate the superior performance of our approach compared with other modified SOTA methods.
\end{itemize}

\section{RELATED WORKS}
Our method draws inspirations from explicit parametrical facial model, synthetic-renderings-based 2D facial avatar and implicit 3D head avatar. So we divide this section into three parts.

\subsection{Explicit Parametrical Facial Model}
Parametric modeling of 3D face has been intensively studied in the past two decades. In the form of explicit meshes, parametric face models are compact, controllable, and easy to be animated. The pioneer work~\cite{CGIT1999Blanz} builds 3D morphable model to represent facial shape, expression, and appearances. Recently, the parametric face models become more expressive by exploiting more powerful modeling techniques, including multi-linear or nonlinear models~\cite{SIGGRAPH2006multilinear, TOG2013sparse, ECCV2014multilinear, Tog2010example, CVPR2018self, CVPR2019towards, CVPR2018nonlinear, CVPR2020LFPF} and the articulated control of expression~\cite{TOG2017flame}.
To model detailed deformations of expression, recent state-of-the-art methods~\cite{TOG2021deca,Danecek2022EMOCA} further learn additional displacement maps with the conditions of image inputs.
Moreover, learning-based generative models such as GAN~\cite{karras2017progressive} or styleGAN~\cite{karras2020a,CVPR2020styleGAN} are also used in existing models~\cite{gecer2021fastganfit,cheng2019meshgan,nagano2018pagan,wang2022faceverse, nagano2019deep, lattas2021avatarme++, luo2021normalized} to enhance the accuracy of facial texture or geometry modeling.
Despite the remarkable progress, all these parametric models can only capture the relatively coarse geometry and appearance of the facial region with the explicit mesh representations, which limits the realism of those reconstruction and animation approaches~\cite{grassal2022neural, cao2016real, hu2017avatar} built upon them.
Instead of solely relying on explicit face models, our approach proposes a controllable hybrid explicit-implicit representation for photo-realistic rendering of 3D face.

\subsection{Synthetic-Renderings-based 2D Facial Avatar}
To utilize the explicit facial model to represent the entire dynamic human head, some methods~\cite{kim2018dvp, koujan2020head2head, doukas2021head2head++, thies2019deferred, thies2020neural} combine classical rendering and learned image synthesis to establish 2D avatar based on the monocular video. Deep Video Portraits~\cite{kim2018dvp} presented impressive full head reenactment and photo-realistic image results based on an image2image translation framework. Head2Head~\cite{koujan2020head2head, doukas2021head2head++} further improved the temporal coherency with a sequential, video-based rendering network. Instead of using the raw texture of the fitted coarse facial model, Deferred Neural Render~\cite{thies2019deferred, thies2020neural} extended the idea by rendering the local feature embedded on the mesh surface. Though the rendering-to-video architecture shows an impressive performance in video portrait synthesis, it does not establish 3D representation for the full-head appearance.

\subsection{Implicit 3D Head Avatar}
In the past three years, it has been an emerging trend to model 3D scenes and objects in an implicit fashion with the success of implicit representations~\cite{mildenhall2020nerf, yariv2020idr}, based on which some works~\cite{chan2022eg3d, wang2022morf, Kellnhofer:2021:nlr, mihajlovic2022keypointnerf} have explored to reconstruct high-fidelity view-consistent 3D appearance for static portraits or ~\cite{park2021hypernerf, park2021nerfies} model the dynamic scene with head movements. As for animatable personalized head character, many methods~\cite{lombardi2019neural, lombardi2021mixture, wang2021learning, gafni2021nerface, grassal2022neural, zheng2021imavatar, athar2022rignerf} attempted to build implicit representation-based personalized full-head avatar. Based on dense multiview capture systems, some researches~\cite{lombardi2019neural, lombardi2021mixture, wang2021learning, cao2021realBinocular} are able to generate facial avatars with impressive subtle details and highly flexible controllability for immersive metric-telepresence. Though the recent work~\cite{cao2022authentic} supports creating authentic avatars from a phone scan, it relies on a prior model that is pretrained in a large-scale multiview-videos dataset captured in a complicated systems. High cost in data acquisition limited the broad applications.
Under light-weight camera settings, based on implicit surface representation, IMAvatar~\cite{zheng2021imavatar} improved generalization to novel expressions by incorporating skinning fields within an implicit morphing-based model, but showed blurry unsatisfying appearance performance. Nerface~\cite{gafni2021nerface} showed state-of-the-art reenactment results with a parameter-controlled neural radiance field, but struggled to extrapolate to unseen expressions. Recently, RigNeRF~\cite{athar2022rignerf} proposed to maintain a canonical neural radiance field with a backward deformation field guided by parametric model mesh, but suffers from the ambiguous correspondences determined by the Euclidean distance.
Besides, for NeRF-based head avatar modeling methods~\cite{gafni2021nerface, hong2021headnerf, guo2021adnerf}, there is a tendency to generate frame-wise inconsistent shape. The problem is originated from the unavoidable noise in the estimation of expressions and head poses, thus the similar input expressions may correspond to slightly-different observed appearances, causing unstable canonical shape recovery. Skillfully incorporating the synthetic renderings of parametric model into neural radiance field, our approach achieves both expressive appearance and robust full-head control, and further addresses the inconsistent shape by modulating feature generation with learnable embeddings.

\begin{figure}
\begin{center}
\includegraphics[scale=0.3]{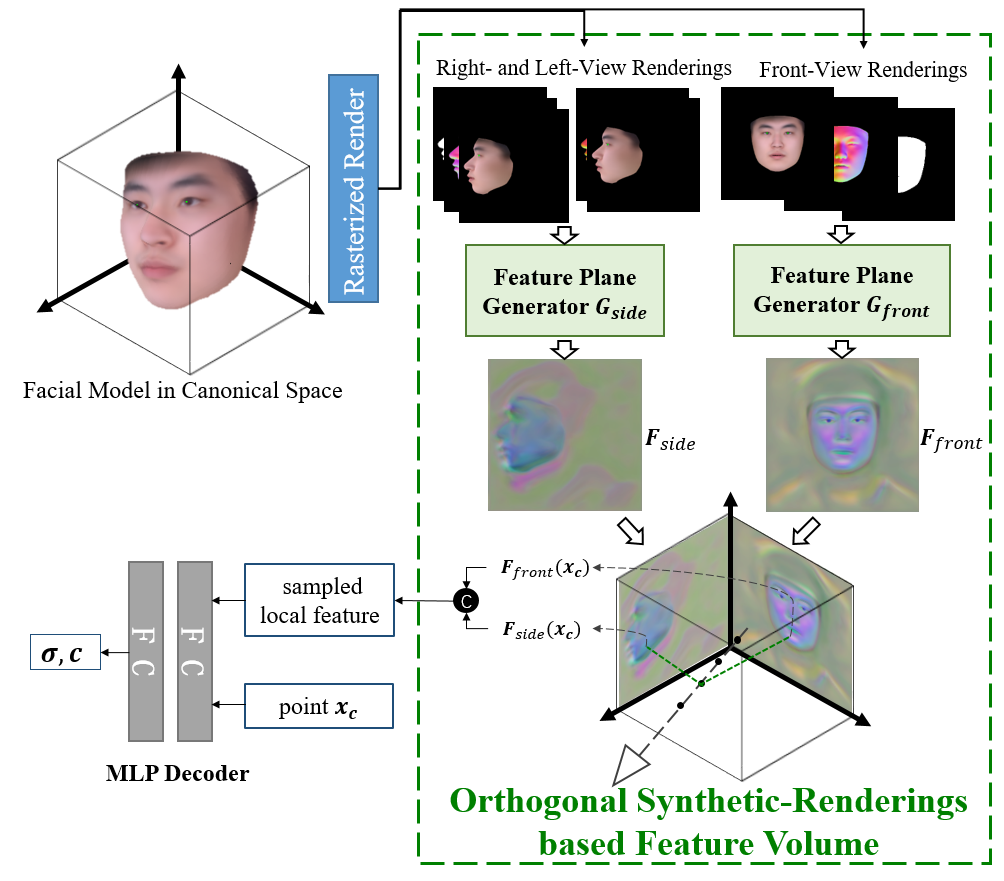}
\end{center}
\caption{Parametric model conditioned Volumetric Representation for canonical head appearance.}
\label{fig:syn-render-nerf}
\end{figure}

\section{Overview}

The overview of our proposed method is illustrated in Fig.~\ref{fig:pipeline}. 
Given the monocular or sparse-view videos, we estimate per-frame facial parametric model $\mathbf{M_t}$ from image sequences $\mathbf{{I}}_t,t=1,\dots,T$. Our method conditions the NeRF on the orthogonal synthetic renderings of the model to describe the expression-related head appearance in the canonical space $H_C$, which supports arbitrary topology and precise expressions control. Besides, per-frame learnable embeddings are utilized to modulate plane feature generation to address expression-shape coupling issue (Sec.~\ref{sec:parametric_model_based_nerf}). Based on the learned LBS weight field, the canonical appearance volume $H_C$ is warped into the observed space $H$ using the estimated head pose, resulting in the decoupled motions of the head and the body (Sec.~\ref{sec:Head_Motion_Decoupling_Module}). With an image-to-image translation network to transferring the volumetrically rendered 2D feature maps to final RGB images, our method achieves high-resolution, photo-realistic and view-consistent portrait image synthesis (Sec.~\ref{sec:Img_Translation}). The overall framework is trained in an adversarial manner with image observations and the established head avatar can be applied for training sequence 4D reconstruction or novel full head reenactment (Sec.~\ref{sec:Optimization_and_Re-Animation}).

\subsection{Recap: Nerface}
\label{sec:Nerface}
Nerface~\cite{gafni2021nerface} firstly extends NeRF~\cite{mildenhall2020nerf} to describe expression-related dynamic head appearance. Based on the classical backbone of 8 fully-connected layers, Nerface additionally inputs low dimensional expressions of the morphable model to condition the neural scene representation network for dynamically changing content. By employing the estimated head pose to transform the rays into the canonical space shared by all frames, the head canonical appearance volume $H_C$ can be formulated as:
\begin{equation}
\label{equ:nerface}
    H_C(\mathbf{x_c}, \mathbf{\gamma_t}, \mathbf{\delta_t}) = (\mathbf{c}, \sigma)
\end{equation}
where the implicit function maps the position in canonical space $\mathbf{x_c}$ to density $\sigma$ and color feature $\mathbf{c}$, under the control of facial expression parameters $\delta_t$, as well as per-frame embeddings $\gamma_t$ to compensate for missing tracking information. 

Nerface relies on global expression blendshape parameters to represent diverse expression-related appearances. However, by simply learning the mapping from the global conditional vectors to appearances with only a short video sequence, it is easy to be overfit. 
Hence, though Nerface is good at faithfully reconstructing the training sequences, without the awareness of the underlying 3D structure of human face, it struggles to generalize to unseen expressions.

\begin{figure}
\begin{center}
\includegraphics[scale=0.2]{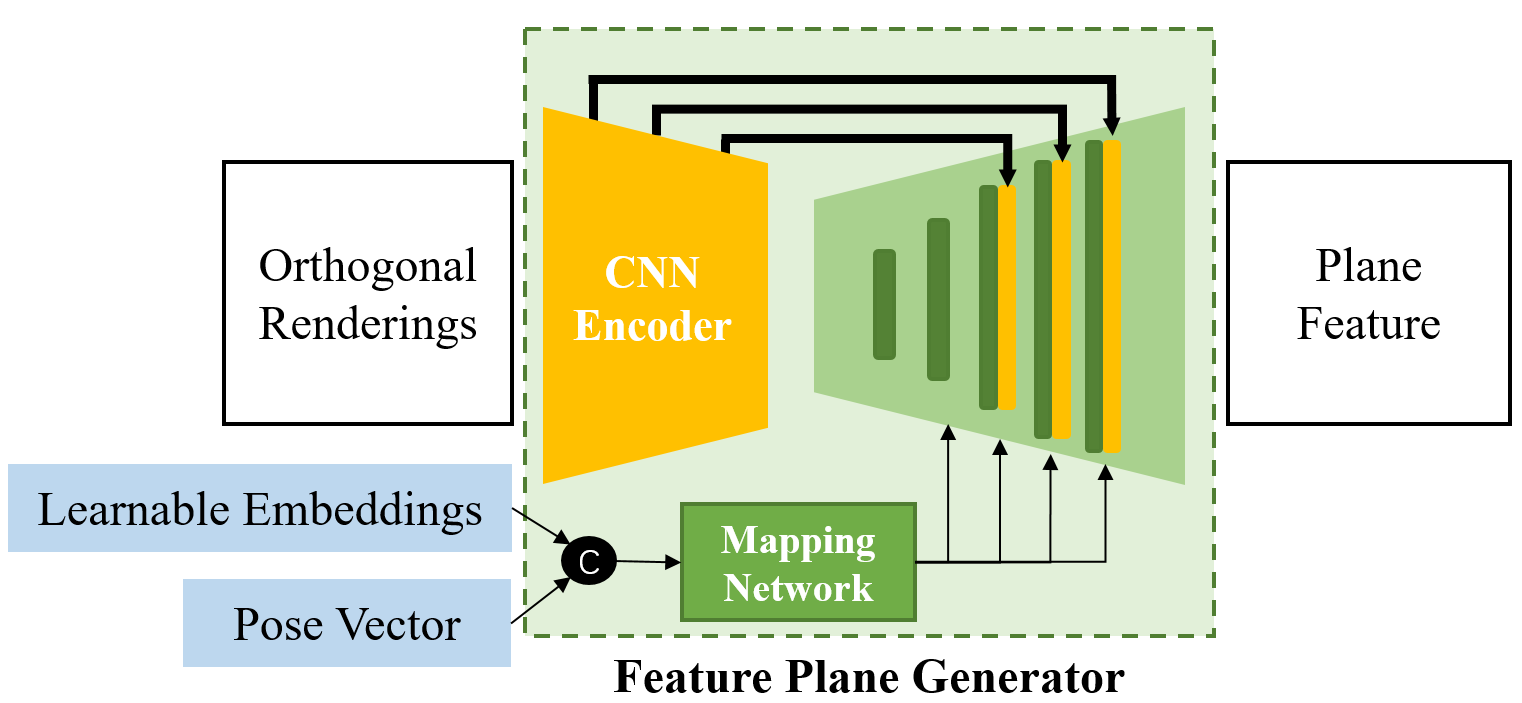}
\end{center}
\caption{The architecture of orthogonal plane feature generator network.}
\label{fig:plane_gen}
\end{figure}

\section{Method}
\subsection{Parametric Model-conditioned NeRF}
\label{sec:parametric_model_based_nerf}
To introduce the facial structure prior into NeRF~\cite{mildenhall2020nerf}, we propose Parametric Face Model-Conditioned Neural Radiance Field. 
Our definition of $H_C$ is reformulated as: 
\begin{equation}
    H_C(\mathbf{x_c}, M_t, \mathbf{\gamma_t}, \mathbf{p_t}) = (\mathbf{c}, \sigma)
\end{equation}
where we utilize the tracked deformed mesh model $M_t$ in zero pose to condition the implicit function, as well as the head pose $\mathbf{p_t}$ to describe the pose-related non-rigid deformation. 

\begin{figure}
\begin{center}
\includegraphics[scale=0.2  ]{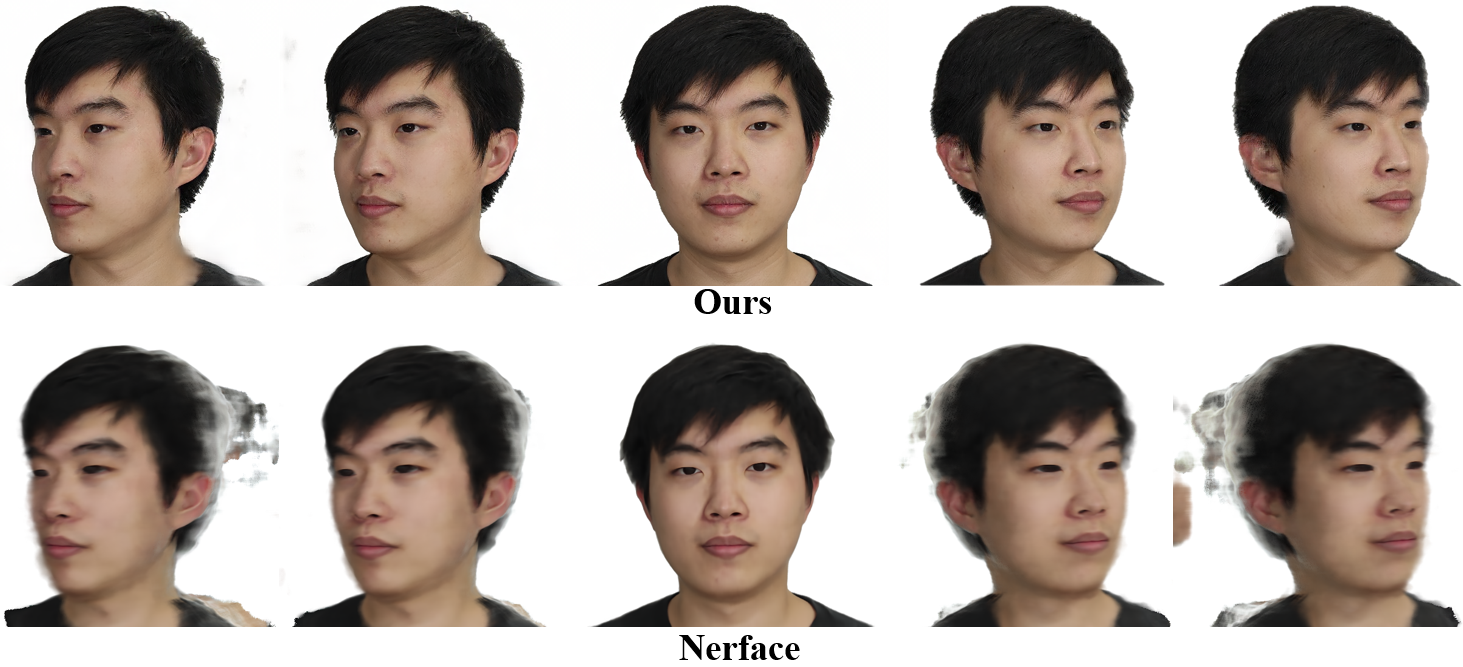}
\end{center}
\caption{We show the novel viewpoint synthesis results of monocular-video-based avatars. By introducing 3D prior into NeRF, our approach improves the robustness of image synthesis under large rotations}
\label{fig:viewConsist}
\end{figure}

\subsubsection{Synthetic-Renderings based Feature Volume}

Fig.~\ref{fig:syn-render-nerf} illustrates the architecture of our NeRF-based representation. The head avatar, embedded with a neural network, is conditioned by model-related local features rather than a global vector for better generalization and precision. Specifically, the orthogonal synthetic renderings of the facial model are leveraged to generate the feature volume for the canonical head appearance.

We orthogonally render the 3D facial model in zero pose and integrate the renderings similar to tri-plane-based neural representation~\cite{chan2022eg3d}.
Considering the special structure of the human head, we abandon the horizontal plane and utilize the front-view and two side-views planes to characterize the head avatar in the canonical space. Instead of sharing one StyleGAN-based backbone to generate all the feature planes, our method utilizes two separate 2D generators to output feature maps individually~\footnote{Removing texture map is also feasible for person-specific avatar modeling, but we empirically find that adding texture renderings can accelerate convergence.}.
It's also feasible to condition one StyleGAN-based backbone with synthetic renderings to generate all feature planes, but we empirically found that utilizing two separate 2D generators individually contributes to accelerate convergence.
As shown in Fig.~\ref{fig:plane_gen}, the synthetic renderings are introduced to the generators to condition the plane feature generation in an explicit manner, for achieving fine-grained controllability. With convolutional encoders extracting image features from the renderings, the extracted multi-resolution features are injected into the generators for spatial-wise feature fusion.
In practice, we generate the front-view plane feature $F_{front}$ based on front-view orthogonal renderings and leverage both left- and right-view renderings to get the side-view plane feature $F_{side}$.
Practically, the deformed mesh $M_t$ is rendered as a normal map, a texture map, and a mask map in each view.
For the experiments reported in this work, each generator produces a $128 \times 128 \times 64$ feature image.

Based on the generated plane feature images, $F_{front}$  and $F_{side}$, for any 3D point in the canonical space, we retrieve its feature vectors via orthogonal projection and bilinear interpolation. All the sampled feature vectors, as well as the positional encoding vector of the coordinate, are concatenated into the point feature $\mathbf{f}$ which is fed into an additional lightweight MLP module with two hidden layers of 128 units. Finally, a scalar density $\sigma$ and a 64-channel color feature $\mathbf{c}$ are predicted for the query point.
Indeed, the combination of orthogonal plane features and light-weight MLP makes the burden of scene representation learning fall on the plane feature generation. Hence, we can rely on the powerful and efficient 2D convolutional network, rather than the large MLP backbone, to extract condition information from synthetic renderings and characterize the dynamic head appearance.

As shown in Fig.~\ref{fig:viewConsist}, with the 3D hint from the facial model, our representation improves the quality of view-consistent image synthesis. The usage of both front, left and right elevation is a succinct but efficient description for 3D human head, containing the full observation of the primary part of the head, as well as getting rid of the constrain from the coarse geometry of the mesh model. Setting more planes will lead to information redundancy and unnecessary memory consumption.

\begin{figure}
\begin{center}
\includegraphics[scale=0.16]{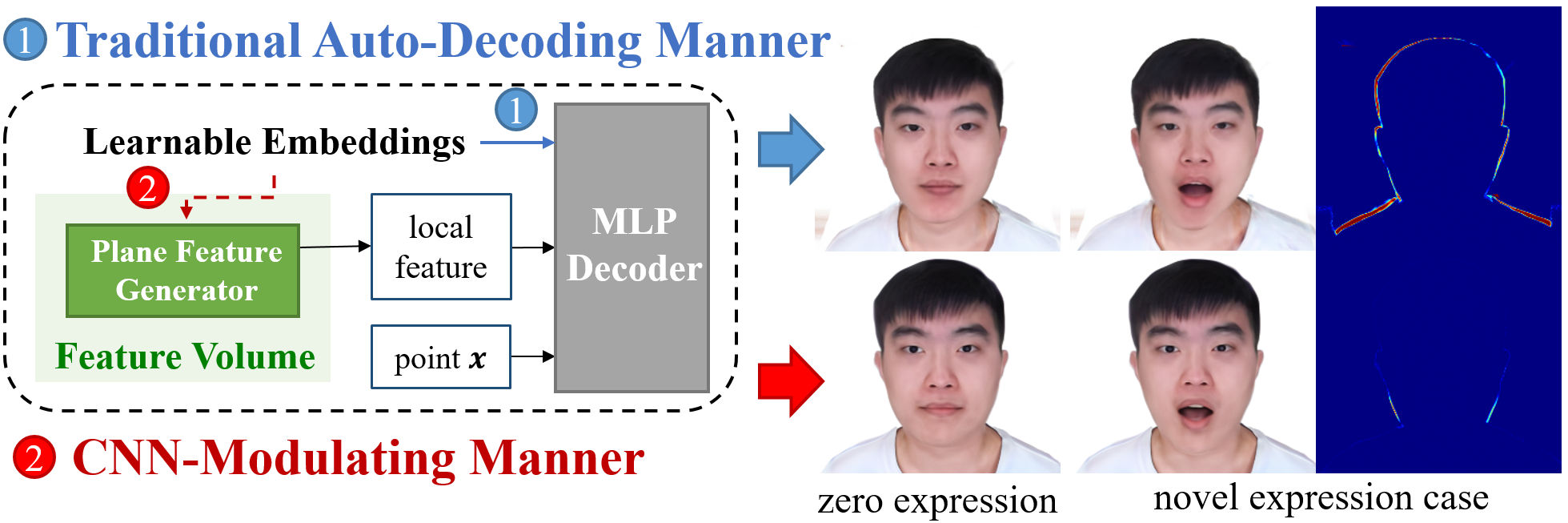}
\end{center}
\caption{Comparison of the two different conditional embeddings. Different from conditioning the learnable embeddings in an auto-decoding fashion (marked as \ding{172}), we utilize them to modulate the generators of the orthogonal plane features (marked as \ding{173}) and prevent embeddings from overfitting to the training dataset. The middle column shows the canonical appearance of the avatar. By only changing the expression (the rightmost column), we illustrate the corresponding rendered appearance and the error map of the generated mask between target cases and the base case.}
\label{fig:LC conditon coparision}
\end{figure}

\subsubsection{Conditional Learnable Embeddings}
\label{sec:latent_condition}
Though our proposed representation is competent for the generation of expression-related canonical head appearance, there is still an unsolved problem: the misalignment between the tracked facial model and the ground-truth observation, which may lead to the frame-wise inconsistent shape. We tackle it by setting additional conditional embeddings for our representation to distinguish similar expressions at different frames.

To account for the misalignment challenges, the previous method\\~\cite{gafni2021nerface} also provides per-frame learnable embedding to the neural head avatar, which contribute to better training sequences reconstruction but cannot eliminate the 
unnatural head shaking while being driven by time-varying expressions in the test phase. This is because it conditions the MLP backbone with embeddings in an auto-decoding fashion~\cite{park2019deepsdf}, causing the embeddings to overfit the training dataset. 
Thanks to our representation that conditions the scene with orthogonal synthetic renderings, we condition the plane feature generators with the learnable embeddings, which are fed into a mapping network to modulate the convolutional kernels of the networks, in the manner of StyleGAN2~\cite{karras2020stylegan2}. 
The embeddings essentially serve as the normalization of the overall feature and concentrate on maximizing global similarity instead of overfitting per-frame local details during the training. Hence our condition manner contributes to producing a latent space with better interpolation performance and learning a consistent expression-independent head shape.
As shown in the Fig.~\ref{fig:LC conditon coparision}, apart from the reasonable expression-related deformation around the cheek, our animation results hardly present shape shaking, proving that our conditional embeddings are able to improve the animation stability.

Specifically, the per-frame embedding is firstly input to a shared mapping network to yield an intermediate latent code which then modulates the convolutional layers of all the separate generators. By constraining the variance of the learnable embeddings, there is a preference to let the generator mainly rely on the synthetic renderings for prediction, and per-frame embedding is utilized to account for the variability resulting from the tracking error.

\subsubsection{Pose-Related Non-Rigid Deformation}
\label{sec:pose_condition}
Though our solution is able to tackle the skeleton motion of head that will be introduced in next section, there still exists pose-related non-rigid deformation caused by head movements in the canonical space, especially in the neck region. In order to describe this, similar to the tackling of per-frame learnable embedding, the estimated head poses are also fed to the mapping network to condition the avatar generation.

\begin{figure}
\begin{center}
\includegraphics[scale=0.16]{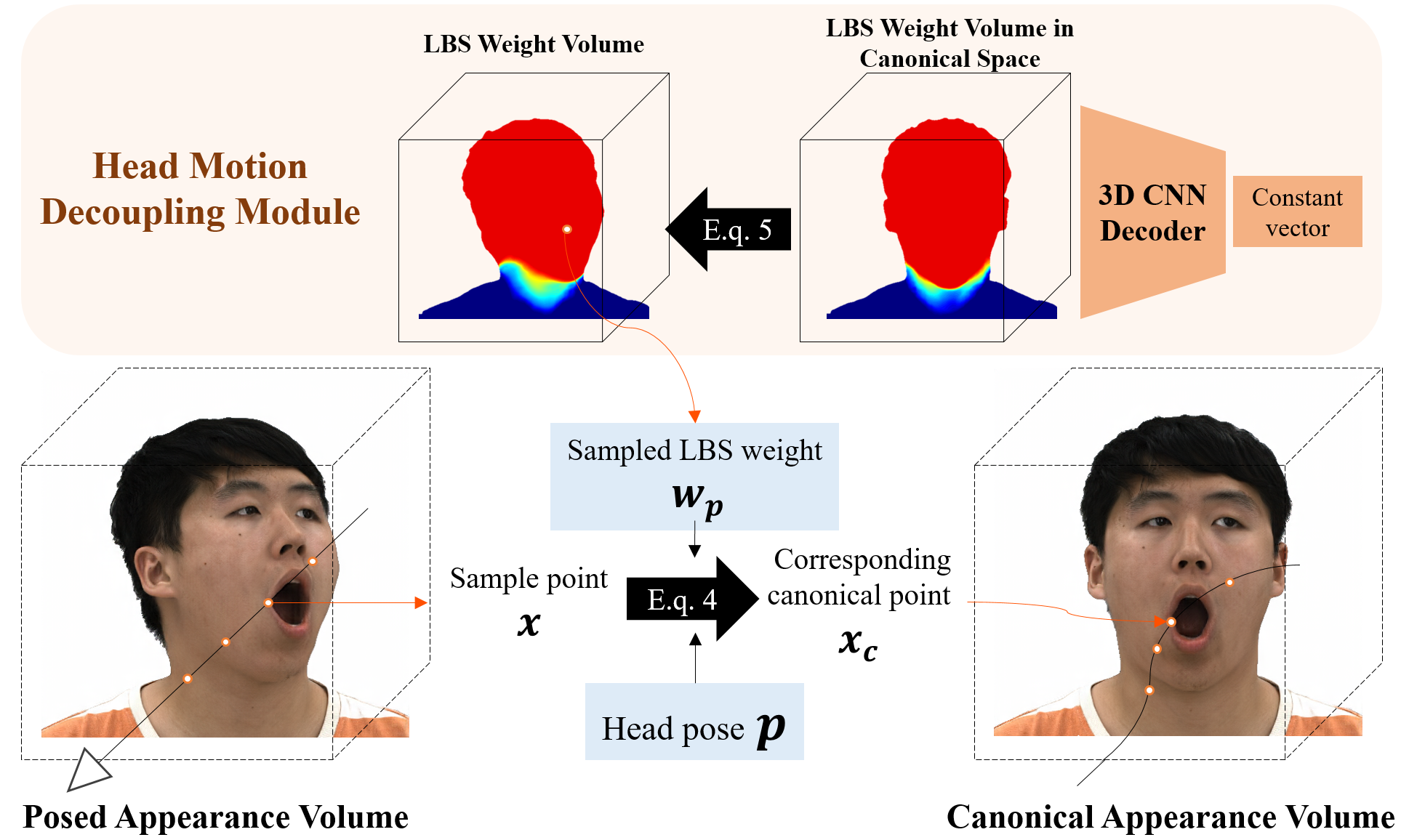}
\end{center}
\caption{Decomposition of head movement. The heatmap in LBS weight volume illustrates that the head (red region) moves according to the pose vector $p$ and the torso (blue region) is hardly affected by the head pose.}
\label{fig:motion_decoup}
\end{figure}

\subsection{Head Motion Decoupling Module}
\label{sec:Head_Motion_Decoupling_Module}

In this section, we will explain how to handle the rigid skeleton deformation driven by head poses. The straightforward treatment in Nerface~\cite{gafni2021nerface}, that the estimated head poses serves as camera poses, leads to the identical motion of both head and body, which is unrealistic. In order to render images agreeing with the ground-truth observation, the relative movement between the head and torso needs to be considered. As shown in Fig.~\ref{fig:motion_decoup}, the canonical appearance volume $H_C$ should be warped to an observed posed appearance volume $H$ with the rigid deformation $T$ :
\begin{equation}
    H(\mathbf{x}, M_t, \mathbf{p_t},  \mathbf{\gamma_t}) = H_C(T(\mathbf{x}, \mathbf{p_t}), M_t, \mathbf{\gamma_t}, \mathbf{p_t})
\end{equation}
Specifically, we compute the head rigid deformation $T$ as inverse linear blend skinning that maps points from the posed space to the shared canonical space:
\begin{equation}
    T(\mathbf{x}, \mathbf{p_t}) = w_p(\mathbf{x})(R_{head}\mathbf{x}+t_{head}) + (1 - w_p(\mathbf{x}))(R_{torso}\mathbf{x}+t_{torso})
\end{equation}
where $w_p$ represents the blend weight, $R_{head}$ and $t_{head}$ the head rotation and translation which comes from estimated head pose $\mathbf{p_t}$, and $R_{torso}$ and $t_{torso}$ means the torso movement which is static by default. In order to avoid overfitting caused by learning backward skinning~\cite{chen2021snarf, zheng2021imavatar}, following HumanNerf~\cite{weng2022humannerf}, we solve for the weight volume in canonical space to derive the $w_p$ as:
\begin{equation}
    w_p(\mathbf{x}) = \frac{w_c(R_{head}\mathbf{x}+t_{head})}{w_c(R_{head}\mathbf{x}+t_{head}) + (1-w_c(\mathbf{x}))} 
\end{equation}
Concretely, we set a 3D convolutional network $W_c$ which inputs a constant random vector and generates the canonical weight volume $w_c(x)$ with limited resolution that can be resampled via trilinear interpolation. With the optimized motion decoupling module, our method can separate out the head movement and stabilize the torso motion.

\subsection{Photo-Realistic 3D-Aware Portrait Synthesis}
\label{sec:Img_Translation}
Although the aforementioned hybrid NeRF-based representation is more expressive than available methods, only relying on pixel supervision ($\mathit{MSE}$/$\mathit{l}_1$ RGB loss) can hardly yield high-frequency details in the rendered images. Hence, we incorporate the 3D representation into an image2image translation architecture and train the overall network jointly in an adversarial manner to enhance facial details and recover realistic portrait images. 

\begin{figure}
\begin{center}
\includegraphics[scale=0.2]{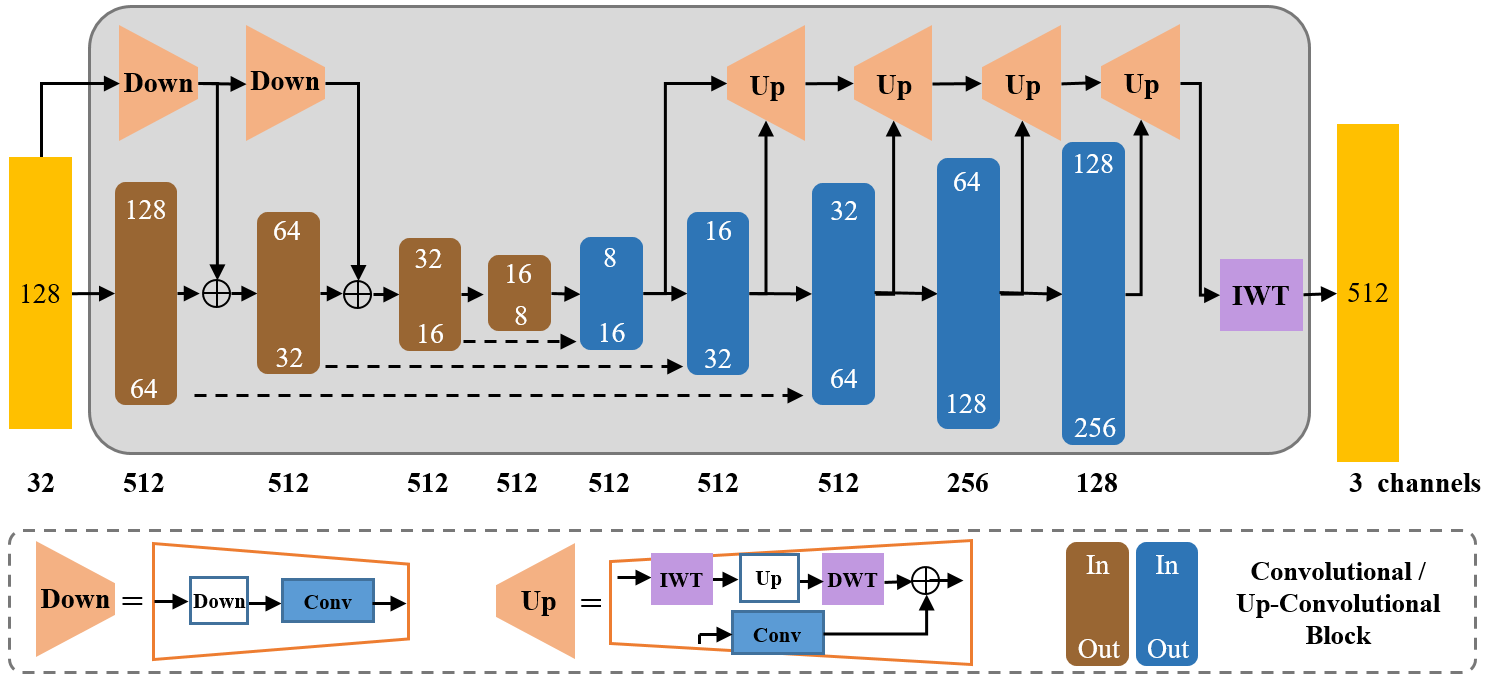}
\end{center}
\caption{The architecture of image2image translation network.}
\label{fig:SRNet}
\end{figure}

Based on the established appearance volume, volume rendering is implemented using two-pass importance sampling as in~\cite{mildenhall2020nerf}.
In order to remain more 3D-aware information for the subsequent module to generate view-consistent images, similar to previous works~\cite{Niemeyer2020GIRAFFE,chan2022eg3d,gu2021stylenerf,hong2021headnerf}, we predict a low-resolution feature map $128 \times 128 \times 64$  from a given camera pose via volumetric rendering, instead of directly rendering an RGB image. 
However, different from these methods leveraging up-sample super-resolution (SR) module, our approach chooses a UNet-style image2image translation network to transfer the raw feature maps to the final RGB images. 
The down-sampling encoding process in UNet helps the 2D network learn the global portrait features, which conduces to the view-consistent images generation. 

Our architecture is presented in Fig.~\ref{fig:SRNet}, which includes two main modifications. First, we incorporate skip connections in the decoder, which map each intermediate feature image to an RGB image and integrate the previous output with the next output through addition. Second, we represent the output image as a wavelet (WT) following~\cite{gal2021swagan}, and the RGB image is generated through an inverse wavelet transform (IWT). This design choice helps reduce the number of parameters and speed up network computations.

The joint training of the overall network can guide NeRF module to provide sufficient and appropriate information for the image2image translation module to raise 3D awareness, for regularizing time- and view- inconsistent tendencies. In the next section, we will explain the training procedure and the used loss functions in detail.

\subsection{Network Training and Avatar Re-Animation}
\label{sec:Optimization_and_Re-Animation}
\subsubsection{Training Strategy}
Given the tracked facial models of the training sequence and segmented mask images, we employ a two-stage training procedure to optimize the neural head avatar, including the pretraining of the NeRF-based appearance volume and the overall joint training.
Firstly, we train only the volume renderer part, the parametric model conditioned NeRF along with the motion decoupling module, to preliminarily establish 3D representation. The objective of the training at the first stage is composed of two components, including an RGB reconstruction loss and a mask loss:
\begin{equation}
    \mathcal{L}_{nerf} = \lambda _{rgb} \mathcal{L}_{rgb}+\lambda _{mask} \mathcal{L}_{mask}
\end{equation}For ease of notation, we drop the subscript (t) of all variables in this subsection.

\textbf{RGB Reconstruction loss:} We additionally set a single linear layer for converting the 64-channel color feature output by the MLP decoder to a 3-channel RGB, and calculate the pixel color via volume rendering~\cite{mildenhall2020nerf}. The main supervision is $\mathcal{L}_{rgb}$ that measures the mean squared error between the rendered and true pixel colors:
\begin{equation}
    \mathcal{L}_{rgb} = \sum_{r\in R} \left \| \hat{C_r} - C(r|M, \mathbf{p},  \mathbf{\gamma} )\right \|  _2^2
\end{equation}
where $R$ is the set of rays in each batch, $\hat{C_r}$ the ground truth pixel color,$C(r | M, \mathbf{p},  \mathbf{\gamma})$ the corresponding reconstructed color determined by parametric model ($M, \mathbf{p} $) and  conditional variables ($\mathbf{\gamma}$) and the network ($H$) via volume rendering function.

\textbf{Silhouette Mask loss:} Additionally, we utilize the foreground mask that can be easily obtained with BgMatting~\cite{BGMv2} algorithm to provide supervision:
\begin{equation}
    \mathcal{L}_{mask} = \sum_{r\in R} BCE(\hat{S_r} , S(r \| \mathbf{\delta}, \mathbf{p},  \mathbf{\gamma}))
\end{equation}
where $BCE(\cdot)$ is the binary cross entropy loss calculated between the rendered silhouette mask value $S(r \| \mathbf{\delta}, \mathbf{p},  \mathbf{\gamma})$ and the ground truth mask $\hat{S_r}$.

Next, we train the whole network end-to-end in an adversarial manner with a discriminator~\cite{gal2021swagan}, using the non-saturating GAN loss~\cite{goodfellow2014gan} with R1 regularization~\cite{mescheder2018training}, denoted $\mathcal{L}_{adv}$. On top of that, the additional loss terms, an $\mathit{l}_{1}$-norm reproduction loss $\mathcal{L}_{recon}$ and a perceptual loss $\mathcal{L}_{percep}$, are utilized to penalize the distance between the synthesized image and the ground-truth image.
\begin{equation}
    \mathcal{L}_{total} = \lambda _{recon} \mathcal{L}_{recon}+\lambda _{percep} \mathcal{L}_{percep}+\lambda _{adv} \mathcal{L}_{adv}
\end{equation}

\begin{figure}
\begin{center}
\includegraphics[scale=0.3]{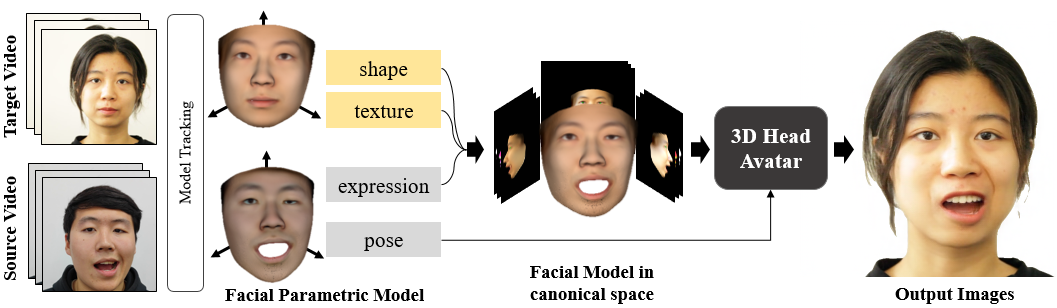}
\end{center}
\caption{Re-animation. For an established head avatar, we implement re-animation by transferring the pose and expression parameters from a facial model estimated from a source video to the avatar facial model.}
\label{fig:reanimation}
\end{figure}

\subsubsection{Full Head Re-Animation}
After network training, the neural head avatar is obtained and can be used to faithfully reconstruct the 4D training sequence and be observed under novel viewpoints. As shown in Fig.~\ref{fig:reanimation}, facial reenactment can be achieved by transferring expression and pose information from the actor to the avatar. Specifically, given a monocular source video, we only need to extract pose and expression parameters from the estimated parametric model for each frame and combine these parameters with our pre-established avatar-specific facial model to generate the sequence of deformed mesh models serving as the network input. As for the conditional embedding vectors, we use the average of all learned embeddings and fix it during the test phase. Finally, the photo-realistic head appearance, which shares the same identity with the modeled avatar but has the novel poses and expressions from the actor in the source video, is generated.

\subsubsection{Implementation Details}
We use Adam optimizer to train our networks with the learning rate to $1\times10^{-3}$ for the image-to-image translation module and $5\times10^{-4}$ for all the others. We use 80 samples (64 from coarse sampling and 16 from fine sampling) per ray. The first stage of training takes about 12 hours and the joint training takes about 36 hours using two NVIDIA 3090 GPUs, while rendering an color image with resolution of 512×512 typically takes 0.15 seconds on one NVIDIA 3090 GPU. 
\begin{figure*}
\begin{center}
\includegraphics[scale=0.43]{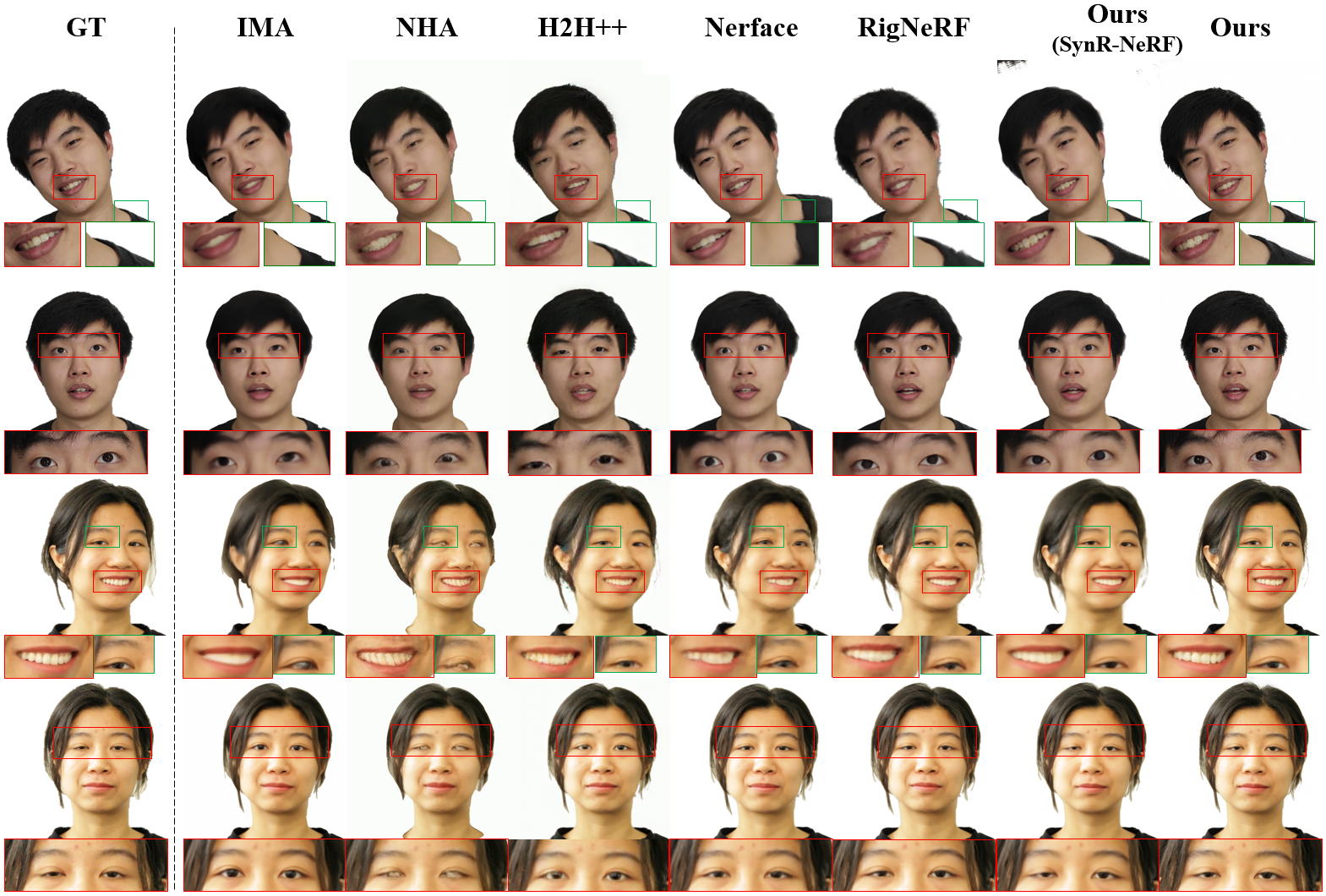}
\end{center}
\caption{Comparison with the state-of-the-art methods on monocular video datasets. From left to right: ground truth images, I M Avatar~\cite{zheng2021imavatar}, Neural Head Avatar~\cite{grassal2022neural},
Head2Head++~\cite{doukas2021head2head++}, Nerface:~\cite{gafni2021nerface}, RigNeRF~\cite{athar2022rignerf}, our NeRF-baseline and ours. The results demonstrate the superior performance of our method in terms of realistic appearance recovery and fine-grained expression control.}
\label{fig:comp_sv}
\end{figure*}

\begin{figure*}[t]
\begin{center}
\includegraphics[scale=0.4]{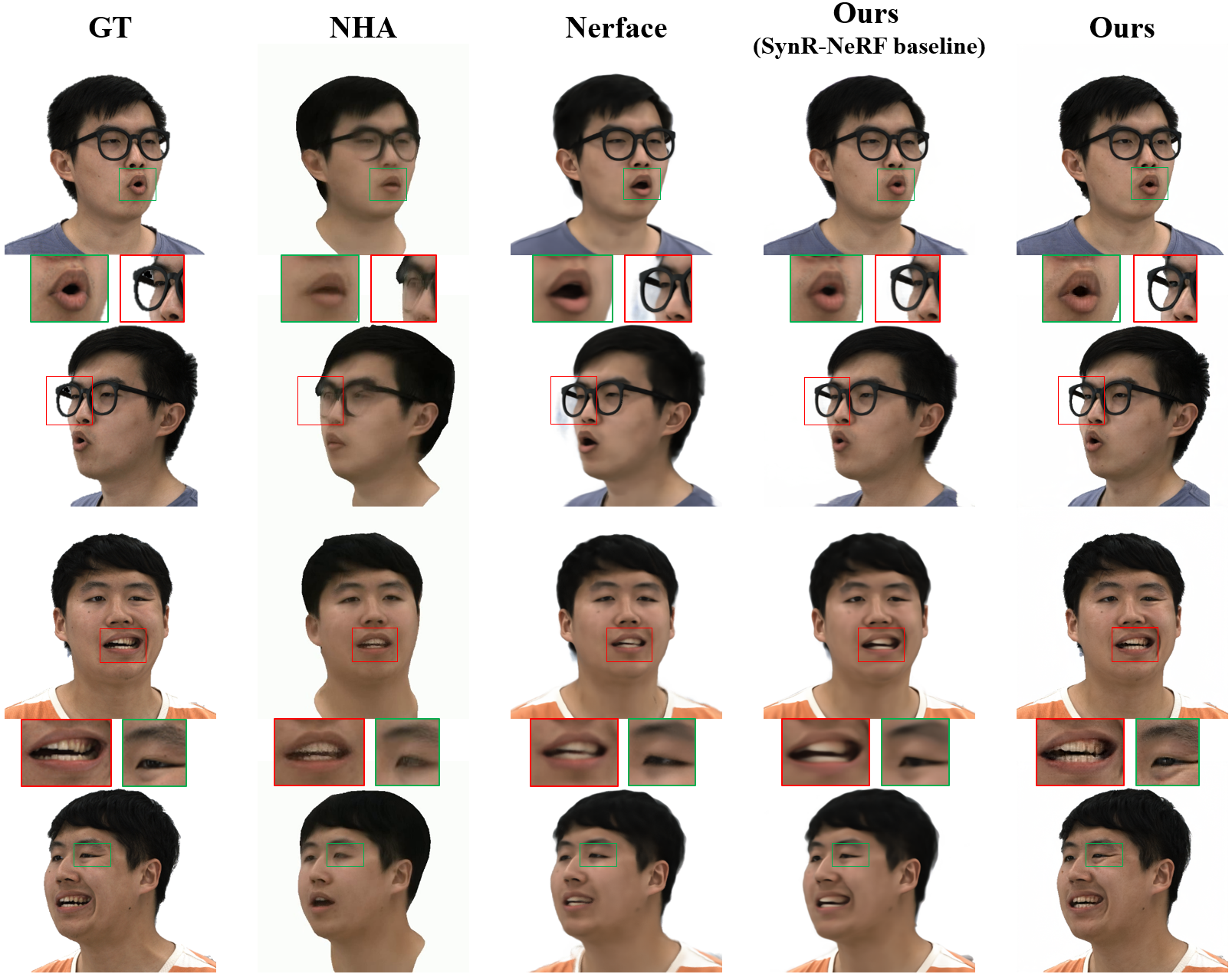}
\end{center}
\caption{Comparison with other methods on multi-view video datasets. From left to righ: ground truth images, Neural Head Avatar~\cite{grassal2022neural}, Nerface:~\cite{gafni2021nerface}, our NeRF-baseline and ours. The results prove our ability in representing topology-varying objects (glasses) and recovering view-consistent high-fidelity appearance.}
\label{fig:comp_mv}
\end{figure*}

\section{EXPERIMENTS}
\textbf{Dataset and Metrics.} 
We separate the evaluation and comparison into two parts: monocular-video-based and multi-view-videos-based experiments.
Our monocular dataset contains the public sequence from I M Avatar~\cite{zheng2021imavatar} and a self-made sequence captured with a phone. We collect multi-view sequences with six cameras focusing on the frontal face. All images are cropped and scaled to 512x512. We calculate the foreground masks with BgMatting~\cite{BGMv2} and estimate th per-frame parametric facial model FaceVerse~\cite{wang2022faceverse} using thir released code. Not that we also track eye gaze and additionally draw the position of pupils on top of the RGB renderings. With each sequence split into training frames and testing ones, we train the networks using the training frames from all viewpoints, and test the animation quality using the testing frames. For quantitative evaluation, we use two standard metrics: peak signal-to-noise ratio (PSNR) and learned perceptual image patch similarity (LPIPS).

\subsection{Comparisons}
\label{subsec:comparsion}

We mainly compare our method with the state-of-the-art 3D head avatar modeling methods:
Nerface~\cite{gafni2021nerface}, IM Avatar (IMA) ~\cite{zheng2021imavatar}, RigNeRF~\cite{athar2022rignerf} and Neural Head Avatar (NHA)~\cite{grassal2022neural}. For the monocular settings, we also compare with 2D facial reenactment method Head2Head++ (H2H++)~\cite{doukas2021head2head++}. We conduct the comparison on the dataset of ~\cite{zheng2021imavatar} and our own data. For IMA, NHA and H2H++, the released data preprocessing codes are utilized to process the monocular videos, and we use our tracked data for Nerface.~\footnote{For fair comparisions, we additionally take the position of pupils besides expression parameters as input.} As the code of RigNeRF is not open-source, we re-implement it and leverage the tracked data of IMA's preprocessing codes to train. To validate the expressiveness of our synthetic-rendering-based NeRF, we also provide a NeRF-baseline(SynR-NeRF) without the image2image translation module.

\begin{figure*}
\begin{center}
\includegraphics[scale=0.4]{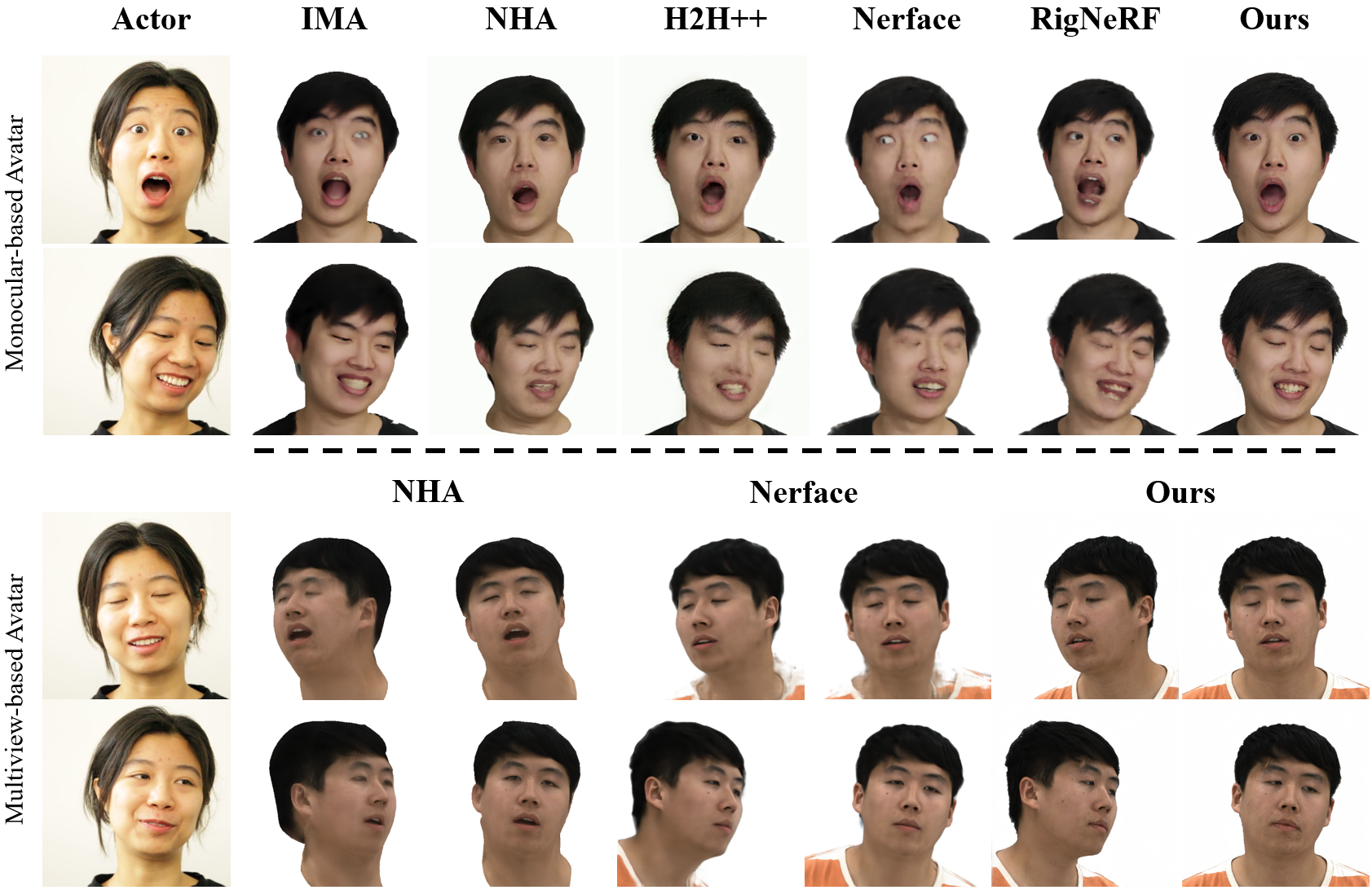}
\end{center}
\caption{Comparison with other methods for the task of head animation. From left to right: actor images, I M Avatar~\cite{zheng2021imavatar}, Neural Head Avatar~\cite{grassal2022neural},
Head2Head++~\cite{doukas2021head2head++},
Nerface:~\cite{gafni2021nerface}, RigNeRF~\cite{athar2022rignerf} and ours. Results demonstrate the generalization of our method to novel expressions and poses.}
\label{fig:comp_Animation}
\end{figure*}

\begin{table}[!ht]
    \centering
    \begin{tabular}{cccccc}
    \hline
        ~ & \multicolumn{2}{c}{case 1} & ~ & \multicolumn{2}{c}{case 2} ~ \\ \cline{2-3} \cline{5-6}
        Method & PSNR $\uparrow$ & LPIPS $\downarrow$ & ~ & PSNR $\uparrow$ & LPIPS $\downarrow$ \\ \hline
        Nerface & 26.47 & 0.221 & ~ & 24.45 & 0.164 \\ 
        IMA & 25.59 & 0.208 & ~ & 23.90 & 0.166 \\ 
        NHA & 19.54 & 0.154 & ~ & 18.21 & 0.158 \\ 
        RigNerf & 27.12 & 0.202 & ~ & 27.92 & 0.118 \\ 
        H2H++ & 24.39 & 0.258 & ~ & 27.12 & 0.154 \\ 
        Ours(SynR-NeRF) & 27.24 & 0.125 & ~ & \textbf{28.78} & 0.109 \\ 
        Ours & \textbf{27.58} & \textbf{0.070} & ~ & 28.476 & \textbf{0.058} \\ \hline
    \end{tabular}
\caption{Quantitative Evaluation for monocular-videos datasets. Case 1 refers to the top two rows of the Fig.~\ref{fig:comp_sv} and case 2 refers to the bottom two rows.}
\label{table:comp_sv}
\end{table}

\begin{table}[!ht]
    \centering
    \begin{tabular}{cccccc}
    \hline 
        ~ & \multicolumn{2}{c}{case 1} & ~ & \multicolumn{2}{c}{case 2} ~ \\ \cline{2-3} \cline{5-6}
        Method & PSNR $\uparrow$ & LPIPS $\downarrow$ & ~ & PSNR $\uparrow$ & LPIPS $\downarrow$ \\ \hline
        Nerface-MV & 21.77 & 0.239 & ~ & 19.90 & 0.247 \\ 
        NHA-MV & 16.39 & 0.238 & ~ & 14.96 & 0.216 \\ 
        Ours(SynR-NeRF) & 22.66 & 0.122 & ~ & 20.06 & 0.15 \\ 
        Ours & \textbf{23.83} & \textbf{0.078} & ~ & \textbf{21.65} & \textbf{0.095} \\ \hline
    \end{tabular}
\caption{Quantitative Evaluation for sparse-views-videos datasets. Case 1 refers to the top two rows of the Fig.~\ref{fig:comp_mv} and case 2 refers to the bottom two rows.}
\label{table:comp_mv}
\end{table}

Qualitative results are presented in Fig.~\ref{fig:comp_sv}. For IMA and NHA, their texture heavily relies on shape reconstruction and the performance of appearance recovery is inferior. Nerface cannot split the head motion and is more prone to generate unstable results for unseen expressions as it lacks structure prior from the parametric model. 
H2H++ suffers from unrealistic image artifacts especially when dealing with challenging head poses. As RigNeRF is built on a backward deformation field guided by a coarse 3DMM mesh, for the unconstrained area such as the mouth interior, RigNeRF tends to establish ambiguous correspondences and generates blurry appearance. Compared with the above approaches, our SynR-NeRF baseline is capable of full-head control and accurate reconstruction of the expressions and head poses. Our full pipeline can moreover recover high-frequency details. The quantitative results presented in Tab.~\ref{table:comp_sv} further demonstrates the superiority of our method. Note that instead of focusing on the pixel-wise similarity, our full pipeline further improves the strength in detail generation and increase the perceptual similarity, which is proven by the gap of LPIPS scores between our method and the other methods.
We also illustrate the comparison on monocular-based animation in Fig.~\ref{fig:comp_Animation}. In this experiment, we utilize part of the video from IMA dataset to drive an established head avatar. While dealing with novel expressions and poses obviously different from the training dataset, our approach shows significantly superior performance and robustness.

For the multi-view settings, 
as far as we know, there is no method focusing on sparse-views-based head avatar modeling available. To this end, similar to our extension to multiview scenario, we extend Nerface-MV and NHA-MV, by adopting multiview parametric face model tracking and optimizing the avatar according to multi-view image evidence.
\footnote{As IMA relies on DECA~\cite{DECA:Siggraph2021} for tracking, which cannot straightforwardly accommodate to multi-view setting, we do not include it in multi-view experiments.} Compared with monocular data, multi-view observations can help model a more complete 3D head avatar, but also cause more obvious misalignment between the estimated mesh models and images due to the limited expressiveness of the parametric model, which raises more challenges for high-quality appearance recovery. Fig. ~\ref{fig:comp_mv} illustrates the qualitative results of two different views, which demonstrates that our method can achieve fine-grained expression control and generate a view-consistent appearance. Nerface tends to produce view-inconsistent artifacts and NHA struggles to describe the topology-varying parts like glasses. The numeric results in Tab~\ref{table:comp_mv}
show that our method achieves higher accuracy in both metrics.
We present the monocular-based animation results in Fig.~\ref{fig:comp_Animation}, which demonstrates our better performance on 3D reenactment.

\begin{table}
    \centering
    \renewcommand\arraystretch{2}
    \begin{tabular*}{\linewidth}{c|c|ccc}
    \hline
         \multicolumn{2}{c|}{PSNR} & \makecell[c]{SynR-NeRF \\ (Ours)} &\makecell[c]{ExprPlanes \\ NeRF} & \makecell[c]{ExprMLP \\ NeRF} \\ \hline
         \multirow{2}{*}{\rotatebox{90}{Backbone}} & MLP &  &  & \checkmark  \\ 
                                   & Orthogonal planes & \checkmark & \checkmark &   \\ \hline 
        \multirow{2}{*}{\rotatebox{90}{Condition}} & Expression vector &  & \checkmark & \checkmark  \\   
                                   & Synthetic rendering & \checkmark &  &  \\ \hline
        \multicolumn{2}{c|}{PSNR} & 26.05 & 22.93 & 22.10 \\
        \multicolumn{2}{c|}{LPIPS} & 0.1516 & 0.1683 & 0.1780 \\ \hline
    \end{tabular*}
\caption{Ablation study on our orthogonal synthetic-rendering based volumetric representation.}
\label{table:abl_nerface2ours}
\end{table}

\begin{figure}
\begin{center}
\includegraphics[scale=0.24]{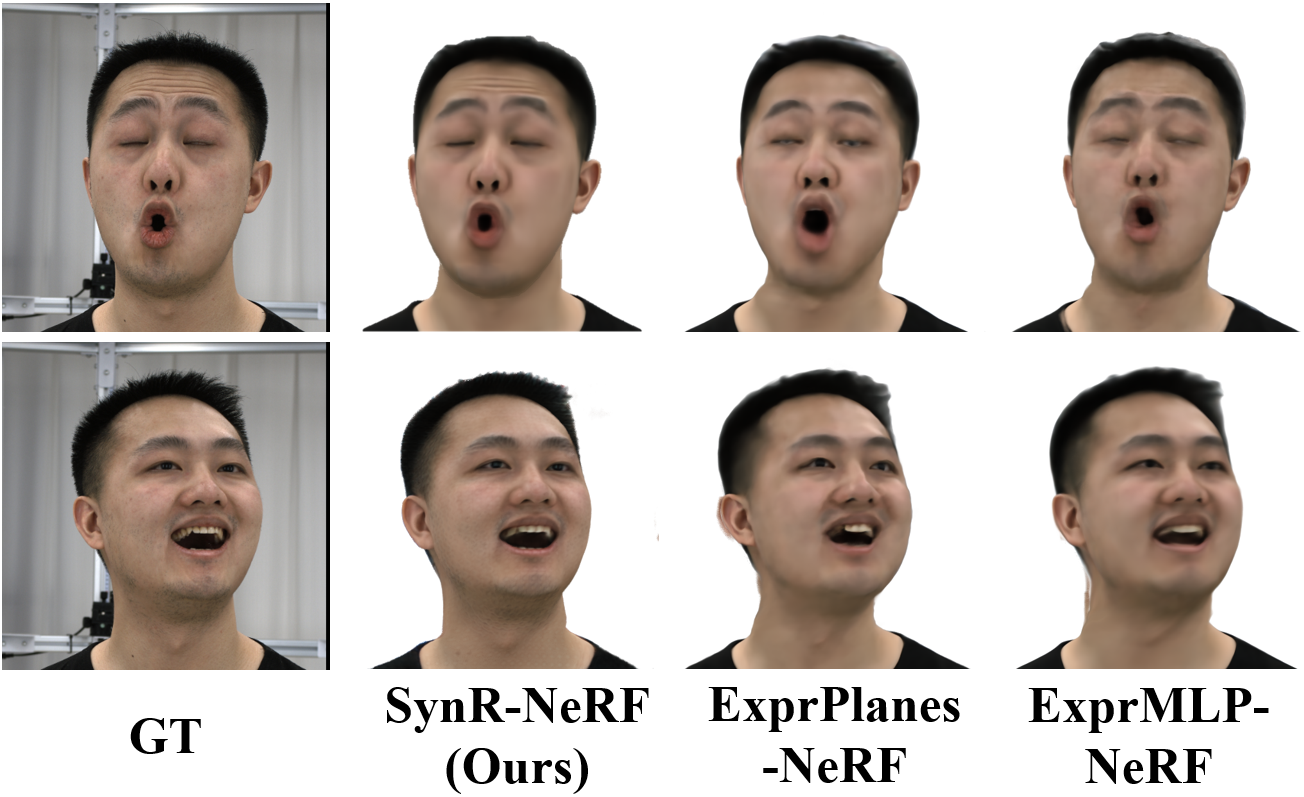}
\end{center}
\caption{Ablation study on our orthogonal synthetic-rendering based volumetric representation.}
\label{fig:nerface2ours}
\end{figure}

\subsection{Ablation Study}

\textbf{Synthetic-rendering based condition}
In this part, two modified baselines were implemented for this ablation study. The first one, named 'ExprPlanes-NeRF', replaces our synthetic-rendering-based condition with the implicit vector-based condition used in Nerface, with all other things being equal. The second baseline, named 'ExprMLP-NeRF', further replaces the orthogonal-planes-based neural representation with a deep MLP backbone used in Nerface. To evaluate the effectiveness of our synthetic-rendering-based volumetric representation, we optimized a head avatar on a monocular video dataset using these two variants, and the results are presented in Fig.~\ref{fig:nerface2ours} and Tab.~\ref{table:abl_nerface2ours}. Comparing 'ExprPlanes-NeRF' and 'ExprMLP-NeRF', we found that only applying the orthogonal-planes representation does not significantly improve performance. However, by using synthetic renderings for explicit condition, our method contributes to more accurate expression control.

\begin{figure}
\begin{center}
\includegraphics[scale=0.17]{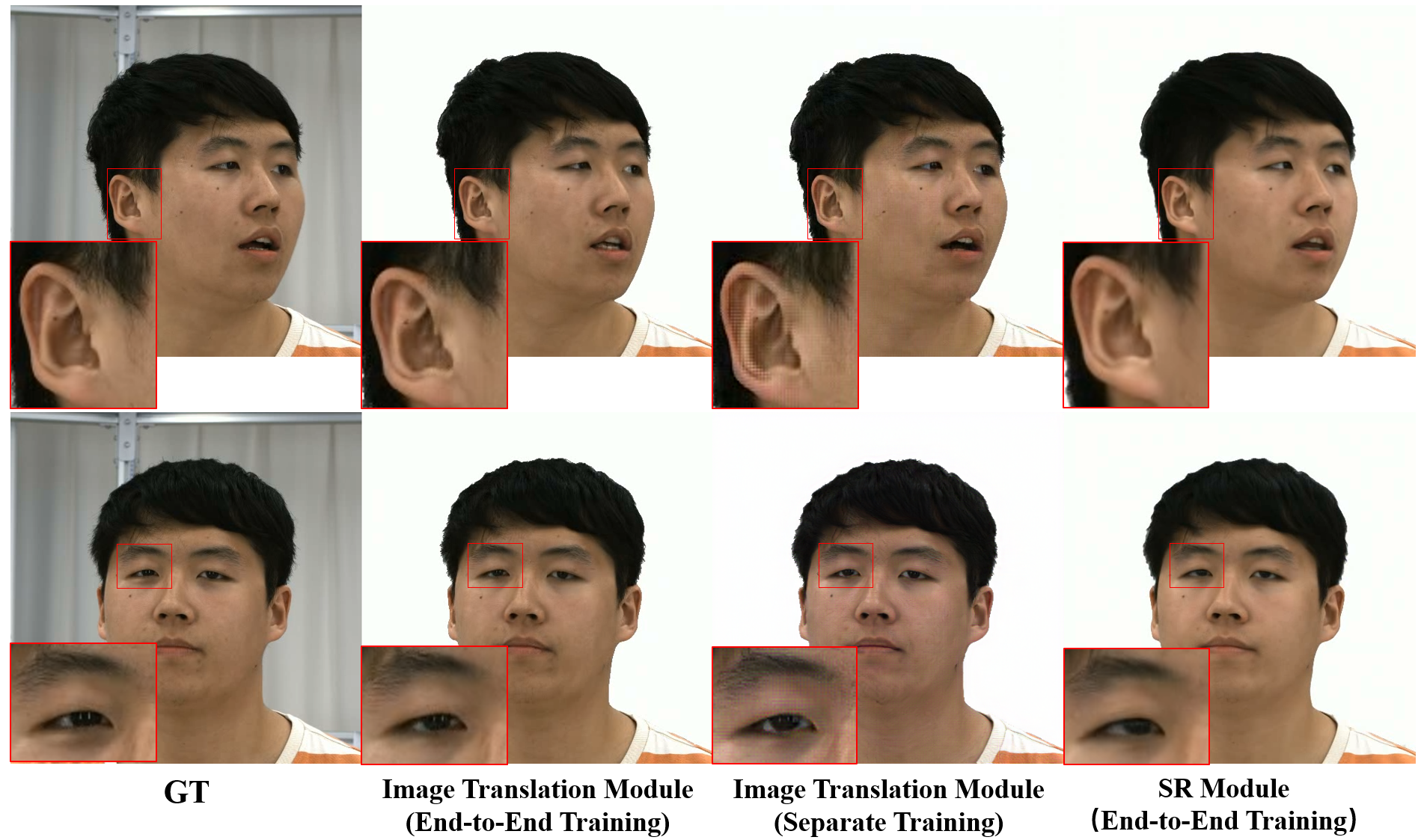}
\end{center}
\caption{Ablation study on the Image Translation Module. Using the image translation module instead of up-sample SR module contributes to recovering fine-scale details and the joint training strategy further helps eliminate image-space artifacts. Please zoom in and also refer to our video for more clear comparisons.}
\label{fig:abl_e2e}
\end{figure}

\textbf{Image-to-Image Translation Module}
Results in sec.~\ref{subsec:comparsion} have proven that the image translation module effectively enhances the fine-level details. 
We implement other two baselines for the ablation study to separately validate the choice of the 2D neural rendering network and the strategy of joint training: 1) We replace the image translation network with the up-sample SR module used by~\cite{chan2022eg3d, Niemeyer2020GIRAFFE} and train the whole pipeline end-to-end. However, when attempting to train the network with adversarial loss functions, we empirically find it hard to maintain stable training. We argue that, without the encoder part, the up-sample SR module alone is not suitable for the person-specific dataset with insufficient diversity. Instead, $\mathit{l}_1 $ loss and perceptual loss are adopted in this experiment. 2) We independently train the image translation module with GAN loss to super-resolution the rendered images from a frozen pretrained SynR-NeRF.
Experiments are conducted on a multi-view sequence, using 5 views for training and leaving one view for evaluation. As shown in Fig.~\ref{fig:abl_e2e} and Tab.~\ref{table:abl_e2e}, up-sample SR based baseline fails to generate fine details. For separate training baseline, it operates primarily in image-space and introduces undesirable inconsistent artifacts, when dealing with the complex distribution of multi-view images. Through end-to-end training the whole framework, our pipeline contributes to guaranteeing realistic detail generation performance.

\begin{table}
    \centering
    \begin{tabular}{c|ccc}
    \hline
        ~ & \textbf{Ours} & \textbf{baseline 1} & \textbf{baseline 2} \\
        ~ & Image Translation  & Up-Sample SR  & Image Translation  \\
        ~ & End-to-End  & End-to-End  & Separate   \\ \hline
        FID & 7.67 & 8.62 & 17.34 \\ \hline
    \end{tabular}
\caption{Ablation study on Image Translation Module.}
\label{table:abl_e2e}
\end{table}

\begin{figure}
\begin{center}
\includegraphics[scale=0.28]{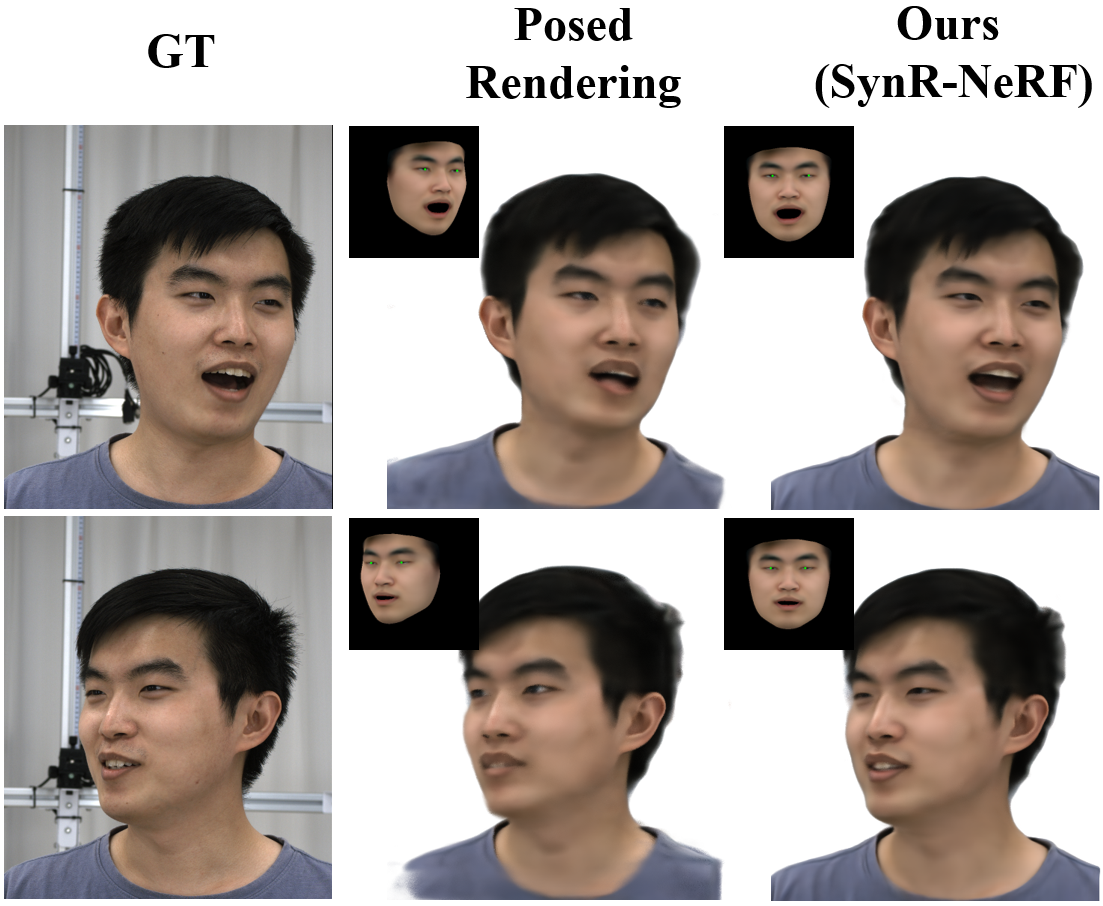}
\end{center}
\caption{Ablation study on orthogonal mesh rendering. Besides the generated portrait images, we also show the front-view orthogonal renderings in the top left corner of the picture.}
\label{fig:abl_posedRender}
\end{figure}

\begin{table}[!ht]
    \centering
    \begin{tabular}{c|cc}
    \hline
        ~ & Posed Rendering & Ours (SynR-NeRF) \\ \hline
        PSNR  & 27.67  & 29.79 \\
        LPIPS & 0.1324 & 0.1219 \\ \hline
    \end{tabular}
\caption{Ablation study on orthogonal mesh rendering.}
\label{table:abl_posedRender}
\end{table}

\textbf{Zero-posed Orthogonal Mesh Rendering}
In our method, we render the 3D facial model orthogonally to create a canonical feature volume in zero pose for feature conditioning. In this part, we introduce a baseline method called 'Posed Rendering,' which involves rendering the pose-dependent meshes to condition the orthogonal feature planes. The results, as shown in Fig.~\ref{fig:abl_posedRender} and Tab.~\ref{table:abl_posedRender}, indicate that 'Posed Rendering' performs worse on the testing set. We attribute this to the coupling of expressions and poses, which creates false correlations between the facial appearance and the face location in the renderings. Plane feature generators have to remember the potential diverse locations in the input image space, leading to reduced performance. In contrast, our method orthogonally renders zero-posed meshes, enabling the generators to concentrate on extracting expression-related information from the renderings and achieve fine-grained control.

\textbf{Conditional Learnable Embeddings}
One strength of our method is that we solve the expression-shape coupling issue presented in previous NeRF-based head avatar methods. We owe it to our strategy of modulating plane feature generators, as metioned in Sec.~\ref{sec:latent_condition}. For evaluation, similar to Fig.~\ref{fig:LC conditon coparision}, we implement a baseline that the NeRF is conditioned in an auto-decoding fashion by inputting the learnable embeddings into the MLP decoder. To fully explore the expression-pose coupling issues, we fix the head pose and only transfer the expression to animate the avatar. The results are presented in the video. Nerface and our modified baseline both show obvious jitter, while our method illustrates stable animation results and better appearance quality.

\begin{figure}
\begin{center}
\includegraphics[scale=0.25]{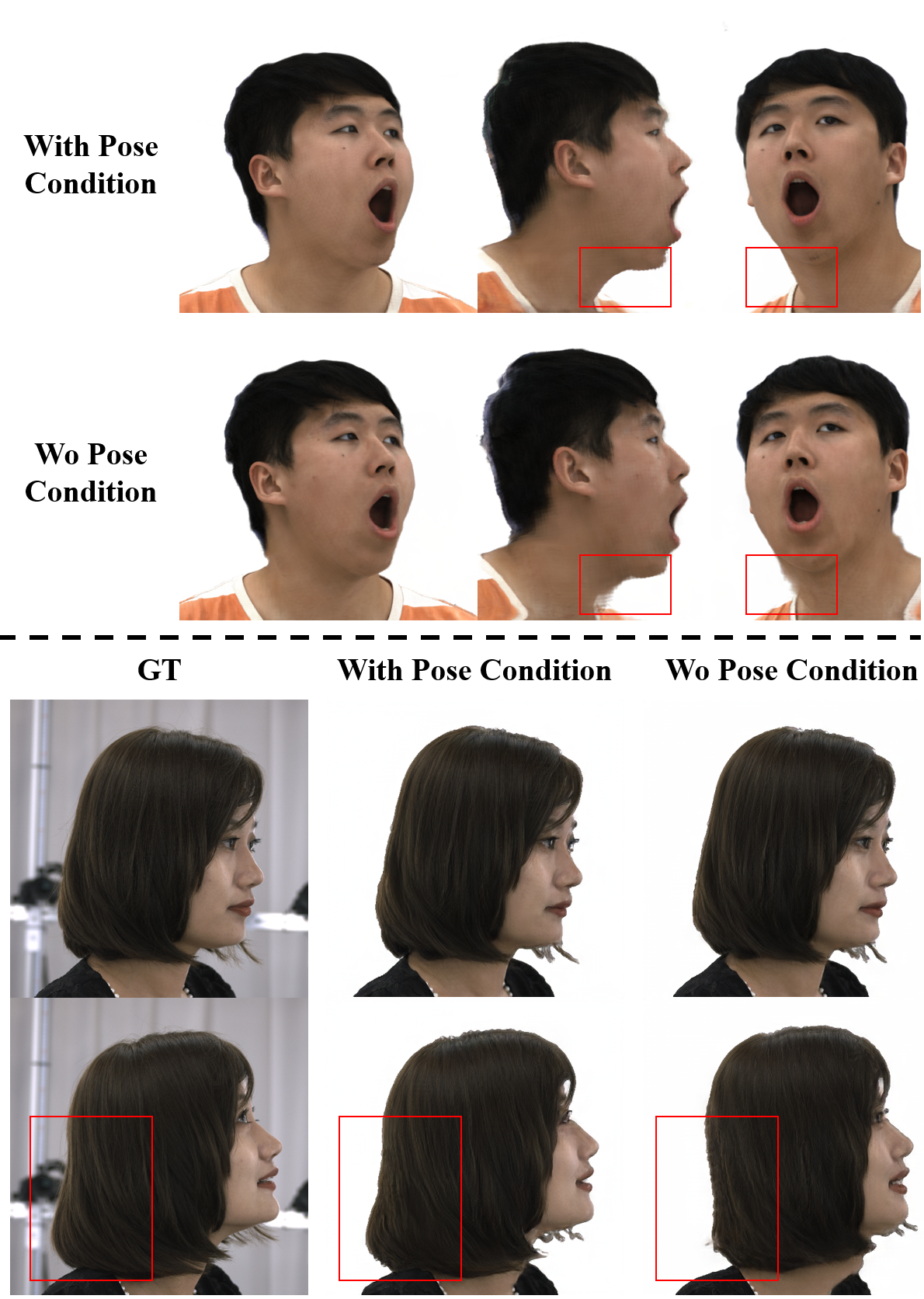}
\end{center}
\caption{Ablation study on pose condition.}
\label{fig:abl_posecond}
\end{figure}

\textbf{Pose Condition}
As mentioned in Sec.~\ref{sec:pose_condition}, by introducing pose vectors into condition, our method is able to describe pose-related non-rigid deformation in the canonical space. Fig.~\ref{fig:abl_posecond} illustrates two cases. 
In the first case, without considering pose, artifacts are visible in the side-view observation when the head turns around. Our method, which includes pose condition, eliminates these artifacts and enhances view-consistency. In the second case, we demonstrate the ability of our method to describe simple pose-related movements of hair.

\section{Discussion and Conclusion}

\begin{figure}
\begin{center}
\includegraphics[scale=0.2]{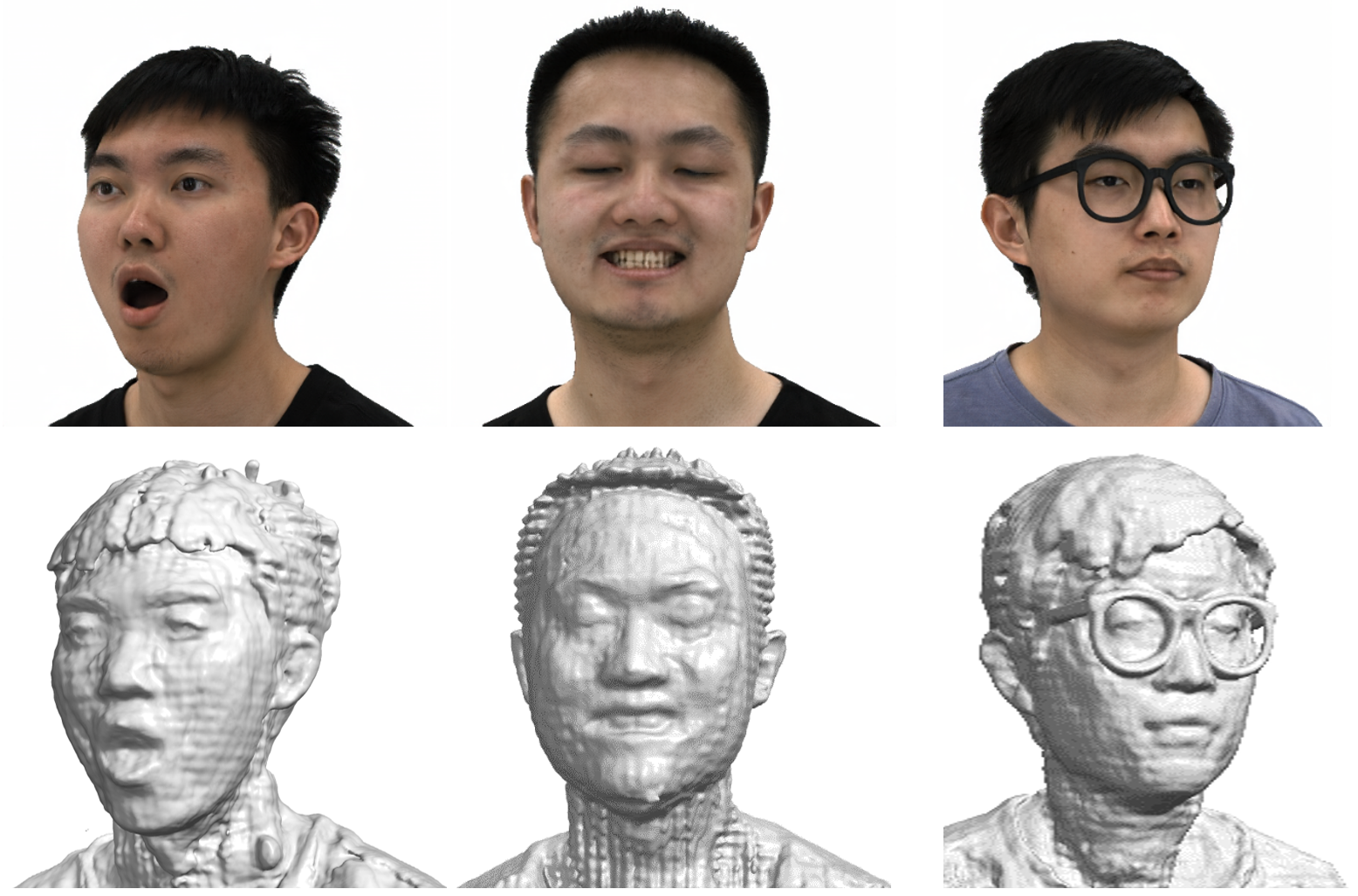}
\end{center}
\caption{The proxy shapes of multiview-based avatars produced by our method. We visualize the 3D shape by using the marching cubes algorithm~\cite{lorensen1987marching} on the density output of our implicit radiance filed to produce a surface mesh.}
\label{fig:limitation}
\end{figure}

\textbf{Limitation.}
Although our approach is able to synthesize high-quality 3D-aware portrait images, the proxy shapes produced by our method cannot be competitive with the state-of-the-art alternative approaches~\cite{zheng2021imavatar, grassal2022neural}, as shown in Fig.~\ref{fig:comp_eg3d} and Fig.~\ref{fig:limitation}.
While this is not important for the photo-realistic stable view-consistent head image synthesis application we consider in this paper, other applications may benefit from reconstructing more accurate morphable geometry.

\begin{figure}
\begin{center}
\includegraphics[scale=0.25]{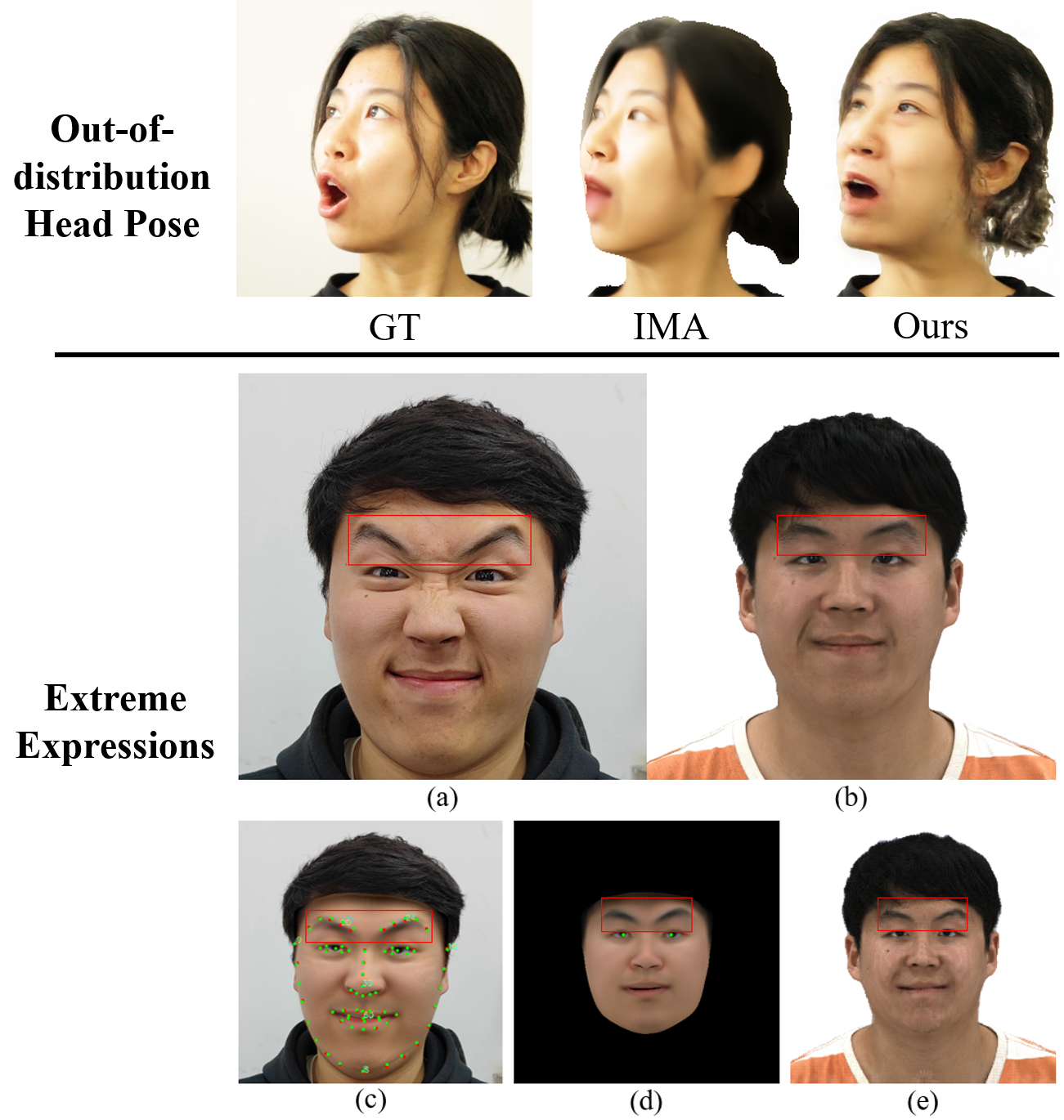}
\end{center}
\caption{Failure cases caused by out-of-distribution head poses and extreme expressions that cannot be expressed by the facial model.
(a) Actor image. (b) Animation result. (c) Estimated facial model and keypoints (red: detected landmarks; green: projected 3D keypoints of the model). (d) Orthogonal front-view synthetic rendering. (e)Synthesized orthogonal front-view result. Note that the eyebrows of our result in (e) align well with the rendering in (d), but the parametric model cannot fully express the narrow eyebrow of the actor image.}
\label{fig:extreme_expr}
\end{figure}

Compared to surface-based avatar modeling methods~\cite{zheng2021imavatar}, our method struggles with out-of-distribution head poses. Additionally, because our method relies on a parametric model to control facial expressions, it is challenging to handle extreme expressions that cannot be expressed by the facial model, as depicted in Fig.~\ref{fig:extreme_expr}. Furthermore, while our method can capture simple pose-related deformation of long hair, it faces difficulties in dealing with challenging topology-varying cases caused by large hair movements.
Special treatment of the hair region is an important problem in the future.

\textbf{Conclusion.}
We introduce a novel modeling method that firstly achieves high-resolution, photo-realistic and view-consistent portrait synthesis for controllable human head avatars. By integrating the parametric face model with the neural radiance field, it has expressive representation power for both topology and appearance, as well as the fine-grained control over head poses and facial expressions. Utilizing learnable embeddings to modulate feature generators, our method further stabilizes animation results. Besides monocular-video-based avatar modeling, we also present high-fidelity head avatar based on a sparse-view capture system. Compared to existing methods, the appearance quality and animation stability of our head avatar is significantly improved.

\begin{acks}
This paper is supported by National Key R\&D Program of China
(2022YFF0902200), the NSFC project No.62125107 and No.61827805.
\end{acks}

\bibliographystyle{ACM-Reference-Format}
\bibliography{sample-acmtog}

\appendix
\section{Comparison with EG3D}

\begin{figure}
\begin{center}
\includegraphics[scale=0.28]{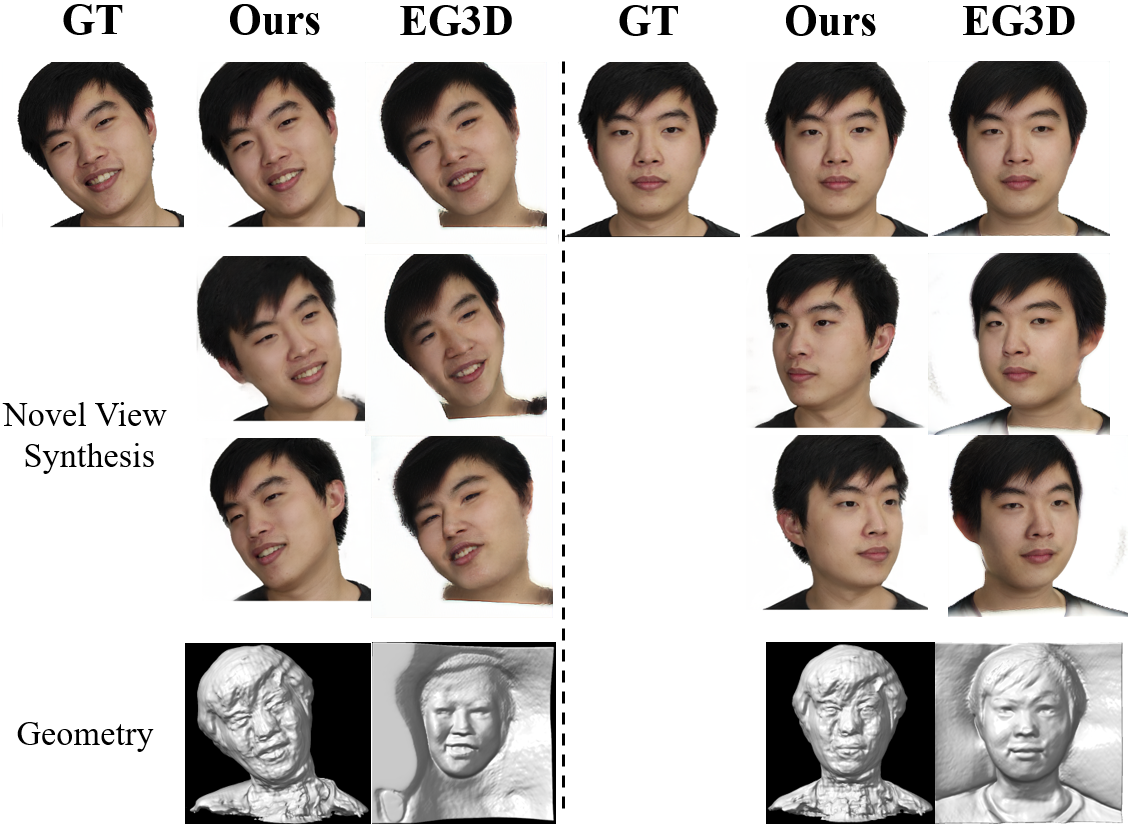}
\end{center}
\caption{Comparison with EG3D on novel view synthesis.}
\label{fig:comp_eg3d}
\end{figure}

EG3D~\cite{chan2022eg3d} is a state-of-the-art powerful generative model for HD 3D portrait, hence we compare it with our method to demonstrate the ability on novel view synthesis. After fitting the pretrained EG3D model with a single reference frame~\cite{roich2021pivotal}, we render the reconstructed 3D head in different views. As Fig.~\ref{fig:comp_eg3d} shows, our method models a more vivid head avatar and presents more convincing novel view synthesis, which benefits from the joint learning of the temporal observation of the person-specific video data.
Besides, EG3D performs worse on poses that are rare in FFHQ dataset.
As for geometry, as we only utilize monocular view observation and do not apply any depth or sigma regularization, the geometry of our head avatar is noisier than EG3D's result.

\begin{figure}
\begin{center}
\includegraphics[scale=0.28]{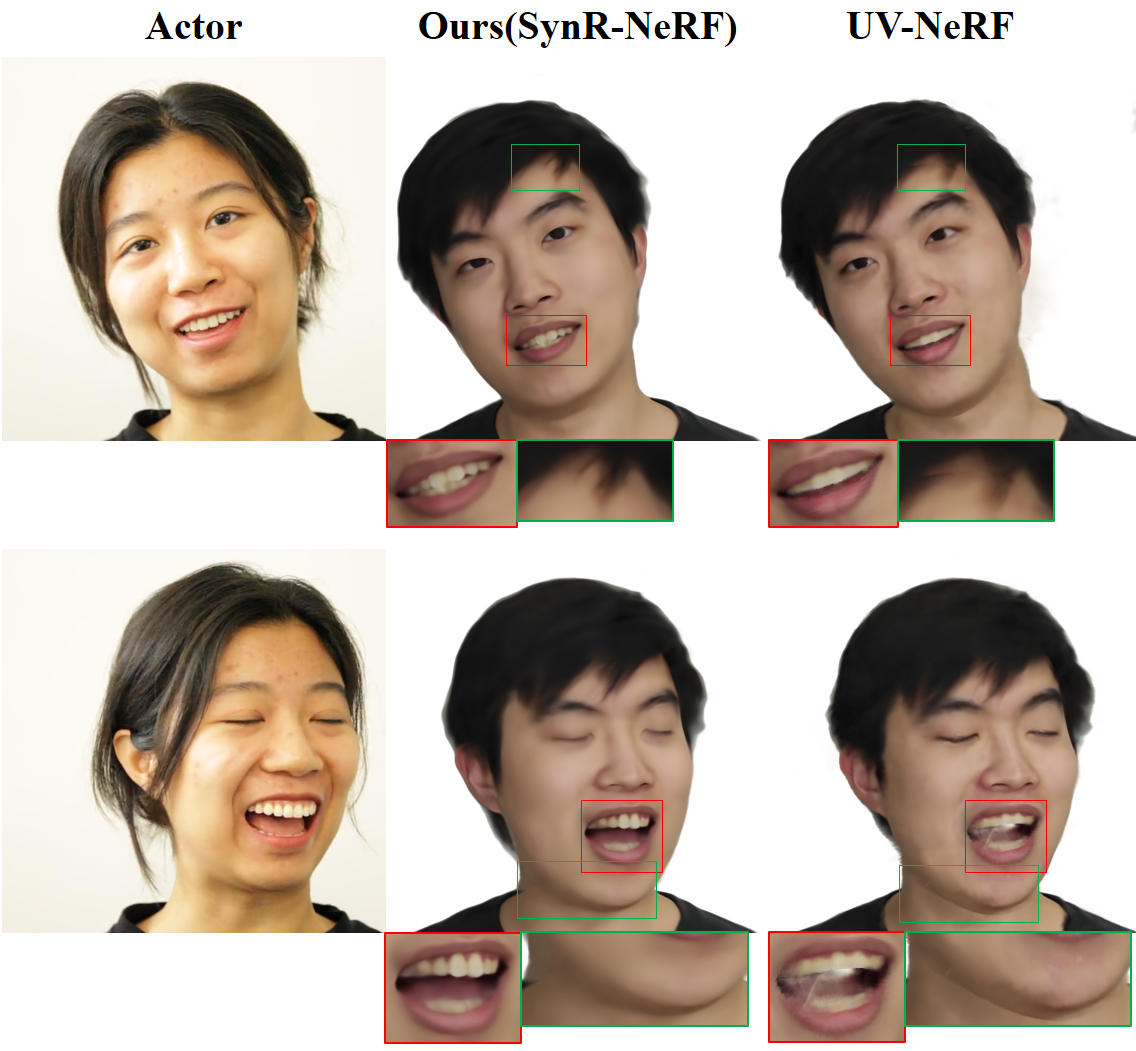}
\end{center}
\caption{Ablation study on model condition manner. Representing the dynamic feature on the mesh surface causes unrealistic artifacts at the edge region of the model mesh. Notice the ambiguous appearance in the mouth and the sharp seam around the neck.}
\label{fig:abl_meshnerf}
\end{figure}

\section{Comparison with UV-based NeRF baseline}
To validate the effectiveness of our adopted synthetic-rendering-based condition, we implement a mesh conditioned baseline (UV-NeRF) that encodes the feature map defined based on the UV parameterization~\footnote{We set a learnable feature map in UV space which is shared for all frames, and utilize a U-Net with 7 layers and the number of channels used are $64, 128, 256, 512, and 512$. For each frame, the U-Net is input with the shared learnable feature map and per-frame UV normal map to generate an expression-related feature map on the UV parameterization. Specifically, the size for our shared UV feature map and generated expression-related UV feature map is $256x256x64$.} instead of our orthogonal plane features. For each sample point, the local feature, obtained from the nearest surface vertices, serves as the input to the MLP decoder. This baseline network is trained under the same setting as our network. The experiment is conducted based on a monocular dataset and the results are presented in Fig.~\ref{fig:abl_meshnerf}. Not surprisingly, though UV-NeRF can accurately reconstruct the expression and reproduce reliable facial appearance, it generates unrealistic artifacts at the edge region of the model mesh. Notice the ambiguous appearance in the mouth and the sharp seam around the neck. Our synthetic-rendering-based condition fully utilizes the powerful convolutional network to learn the reasonable correspondence between the facial model and the entire head appearance, synthesizing consistent and stable images.

\section{Ablation studies on the textured mesh rendering condition}
\begin{figure}
\begin{center}
\includegraphics[scale=0.4]{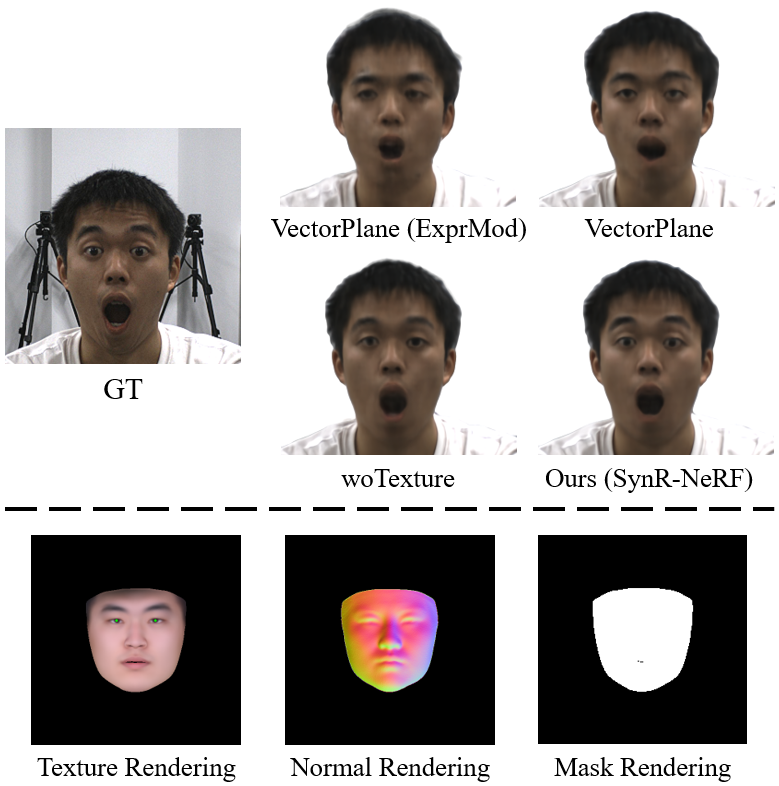}
\end{center}
\caption{Ablation studies on the textured mesh rendering condition. The last row visualizes the synthetic renderings of the front view.}
\label{fig:abl_woTexture}
\end{figure}

To further validate our synthetic-rendering-based condition in detail, we implement two baselines that generate the feature volume directly from latent codes. The first one, called 'VectorPlane', uses the expression parameters as input for the feature generation backbone, while the second one, 'VectorPlane(ExprMod)', feeds the expression parameters to a mapping network to modulate the convolutional kernels of the networks. Both baselines modulate the feature generation backbone with pose vectors and per-frame latent codes. As shown in Fig.~\ref{fig:abl_woTexture} and Tab.~\ref{table:abl_woTexture}, learning the mapping from the global vectors to appearances tends to overfit training sequences, and the ability of expression control degraded for out-of-distribution expressions.

We expound our analysis on the differences between the two conditional methods.
Two similar expressions may be too close in the parameter space for a global-vector-conditioned feature generator to distinguish between them. Our synthetic-rendering-based conditional method preserves the spatial alignment between the mesh renderings and the feature planes, hence the local spatial changes in the rendering image space can be reflected in the feature volume. Additionally, as we align individual local features at the pixel level of renderings to the global context of the entire appearance, it is more likely to infer plausible results for unseen expressions.

Besides, we also implement a baseline, named 'woTexture', that only utilizes normal and mask renderings to condition feature volume generation. Removing texture rendering is feasible for person-specific avatar modeling and does not significantly affect the robustness of expression control. However, despite the similar numerical results, the visualization results of our method exhibit more detailed appearances around the eyes. We hypothesize that the texture rendering contains more high-frequency information around the eye region, as shown in the last row of Fig.~\ref{fig:abl_woTexture}, which may facilitate the network in effectively learning dynamic appearance around eyes.

\begin{table}
    \centering
    \begin{tabular}{c|cccc}
    \hline
        ~ & \makecell[c]{VectorPlane \\ (ExprMod)} & VectorPlane & woTexture & SynR-NeRF \\ \hline
        PSNR  & 26.99 & 27.53  & 28.40 & 30.20 \\
        LPIPS & 0.1446 & 0.1358 & 0.1268 & 0.1243 \\ \hline
    \end{tabular}
\caption{Ablation studies on the textured mesh rendering condition.}
\label{table:abl_woTexture}
\end{table}

\section{Comparison with 2D re-enactment based baseline}
\begin{figure}
\begin{center}
\includegraphics[scale=0.4]{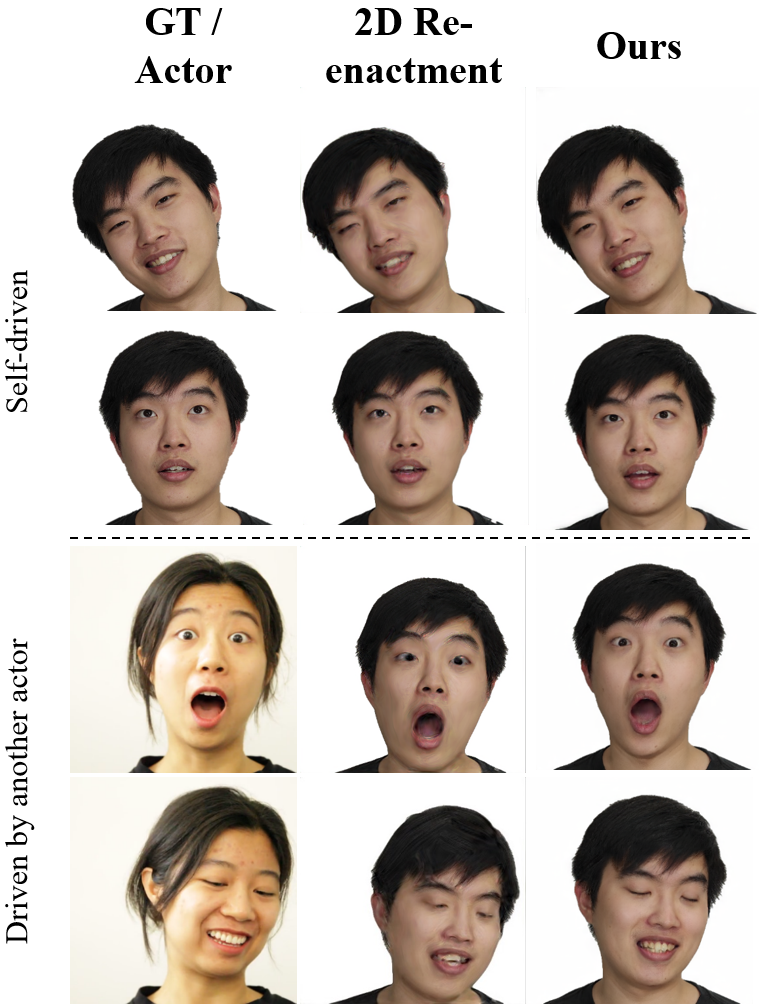}
\end{center}
\caption{Comparison with a 2D re-enactment baseline on monocular video datasets. The results demonstrate the superior performance of our method in terms of realistic appearance recovery and robust expression/pose control.}
\label{fig:abl_2Dreenact}
\end{figure}

By omitting the Nerf module, we implement a 2D re-enactment method that only utilizes our image2image translation module. In our practice, the image2image translation network takes the renderings of the fitted 3DMM in the observed view as input and generates the corresponding 2D images. This 2D re-enactment method has limitations. Firstly, it cannot establish an avatar based on a multi-view dataset because it cannot differentiate between camera poses and head poses. Secondly, when applied to monocular videos, the 2D re-enactment method is sensitive to the location of the facial model in the image space, as shown in Fig.~\ref{fig:abl_2Dreenact}. The 2D re-enactment method is prone to generating artifacts or distorted faces, particularly at the edges of the image.

\section{Multi-view Setting}
While fitting per-frame 3DMM model, we use the detected landmarks in multi-view images at the same instant as supervision, and additionally estimate the scale parameter of 3DMM. As for NeRF optimization, we simply leverage multi-view images of the same frame to supervise the appearance, with all loss terms and the training strategy as same as monocular-based setting. 

\section{Data Preprocessing}
We optimize the shape and texture parameters using the first few frames of the video and these parameters remain fixed for the remaining frames of the video. For each frame's 3DMM fitting, we optimize the pose, expression, and illumination parameters, which are initialized as the last frame's fitting results. Once the 3DMM model is tracked, we utilize PyTorch3D~\cite{ravi2020pytorch3d} to render per-frame synthetic renderings onto orthogonal planes. The tracking code mainly comes from the open-source project Faceverse~\footnote{ (https://github.com/LizhenWangT/FaceVerse)}.
Besides, to perform eye gaze tracking, we segment the dark area within the eye region in the given frame. The centroid of the dark area is considered the pupil, and we calculate the pupil's relative position inside the eye based on the detected landmarks surrounding the eye. Finally, we mark the pupil as a small dot in the front-view orthogonal renderings.

\section{Inference Time}
Tab.~\ref{table:time} shows the detailed time consuming during inference. Rendering a color image with a resolution of 512x512 takes 0.15 seconds on one NVIDIA 3090 GPU, and the most time-intensive part is the volume rendering of our NeRF module.

\begin{table}
    \centering
    \begin{tabular}{|c|c|c|c|c|}
    \hline
        \makecell[c]{Data IO} & \makecell[c]{Feature Plane \\ Generation} & \makecell[c]{NeRF Module} & \makecell[c]{2D Image \\ Module} & \textbf{Total}\\ \hline
        47.4ms & 22.3ms & 58ms & 24.5ms & 152.2ms \\ \hline
    \end{tabular}
\caption{Detailed time consuming during inference.}
\label{table:time}
\end{table}

\end{document}